%% file: _main.tex
\input{_constants}
\arxiv 

\pdfoutput=1
\documentclass[10pt,twocolumn,letterpaper]{article}
\input{cvpr_header}
\begin{document}
\title{\paperTitle}
\author{\authorBlock}

\definecolor{yzybest}{rgb}{0.98, 0.8, 0.8} 
\definecolor{yzysecond}{rgb}{0.99, 0.88, 0.77} 
\definecolor{yzythird}{rgb}{1.0, 1.0, 0.8} 

\newcommand{\bestcolor}{\cellcolor{yzybest}}
\newcommand{\secondcolor}{\cellcolor{yzysecond}}
\newcommand{\thirdcolor}{\cellcolor{yzythird}}

\ifdefined\onlyappendix
{
}
\else
{
\twocolumn[{%
\renewcommand\twocolumn[1][]{#1}%
\maketitle

\input{figs/tex/teaser}

}]
\let\thefootnote\relax
\footnote{
$\dag$: the corresponding author.
}

\input{00_abstract}
\input{01_intro}
\input{02_related}
\input{03_method}
\input{04_experiments}
\input{10_conclusion}
}
\fi

{\small
\bibliographystyle{ieeenat_fullname}
\bibliography{11_references}
}

\ifarxiv \appendix \input{12_appendix} \fi

\ifreview \appendix \input{12_appendix} \fi

\end{document}

%% file: _constants.tex
\def\paperID{XXXXXX}
\def\confName{CVPR}
\def\confYear{2024}

\def\paperTitle{Human101: Training 100+FPS Human Gaussians in 100s from 1 View}

\def\authorBlock{
    Mingwei Li \qquad
    Jiachen Tao \qquad
    Zongxin Yang \qquad
    Yi Yang$^{\dag}$ \\
    ReLER, CCAI, Zhejiang University
}

\newif\ifreview 
\newif\ifarxiv \newcommand{\arxiv}{\arxivtrue}
\newif\ifcamera 
\newif\ifrebuttal 

%% file: cvpr_header.tex
\ifreview \usepackage[review]{cvpr} \fi
\ifarxiv \usepackage[pagenumbers]{cvpr} \fi
\ifrebuttal \usepackage[rebuttal]{cvpr} \fi
\ifcamera \usepackage{cvpr} \fi

\input{_macros}  

\usepackage{xr-hyper}
\usepackage{amsmath}
\usepackage[ruled,vlined]{algorithm2e} 
\usepackage{algorithmic}
\usepackage{amssymb} 
\makeatletter
\newcommand*{\addFileDependency}[1]{
  \typeout{(#1)}
  \@addtofilelist{#1}
  \IfFileExists{#1}{}{\typeout{No file #1.}}
}

\makeatother

\definecolor{cvprblue}{rgb}{0.21,0.49,0.74}
\usepackage[pagebackref,breaklinks,colorlinks,citecolor=cvprblue]{hyperref}
\usepackage[capitalize]{cleveref}
\crefname{section}{Sec.}{Secs.}
\crefname{table}{Table}{Tables}
\crefname{figure}{Fig.}{Figs.}

%% file: _macros.tex
\usepackage{graphicx}	
\usepackage{amsmath}	
\usepackage{amssymb}	
\usepackage{booktabs}
\usepackage{times}
\usepackage{microtype}
\usepackage{epsfig}
\usepackage[table,xcdraw,dvipsnames]{xcolor}
\usepackage{caption}
\usepackage{float}
\usepackage{placeins}
\usepackage{color, colortbl}
\usepackage{stfloats}
\usepackage{enumitem}
\usepackage{tabularx}
\usepackage{xstring}
\usepackage{multirow}
\usepackage{xspace}
\usepackage{url}
\usepackage{subcaption}
\usepackage{xcolor}
\usepackage{esvect}
\usepackage{fontawesome}
\usepackage{makecell}
\usepackage[table]{xcolor}

\usepackage[hang,flushmargin]{footmisc}

\ifcamera \usepackage[accsupp]{axessibility} \fi

% ...

\ifarxiv  \fi

\newcommand{\R}[1]{{%
    \textbf{%
        \ifstrequal{#1}{1}{\textcolor{red}{R#1}}{%
        \ifstrequal{#1}{2}{\textcolor{blue}{R#1}}{%
        \ifstrequal{#1}{3}{\textcolor{magenta}{R#1}}{%
        \ifstrequal{#1}{4}{\textcolor{teal}{R#1}}{%
                           \textcolor{cyan}{R#1}%
        }}}}%
    }%
}}

%% file: figs/tex/teaser.tex
\begin{center}
    \centering
    \includegraphics[width=\linewidth]{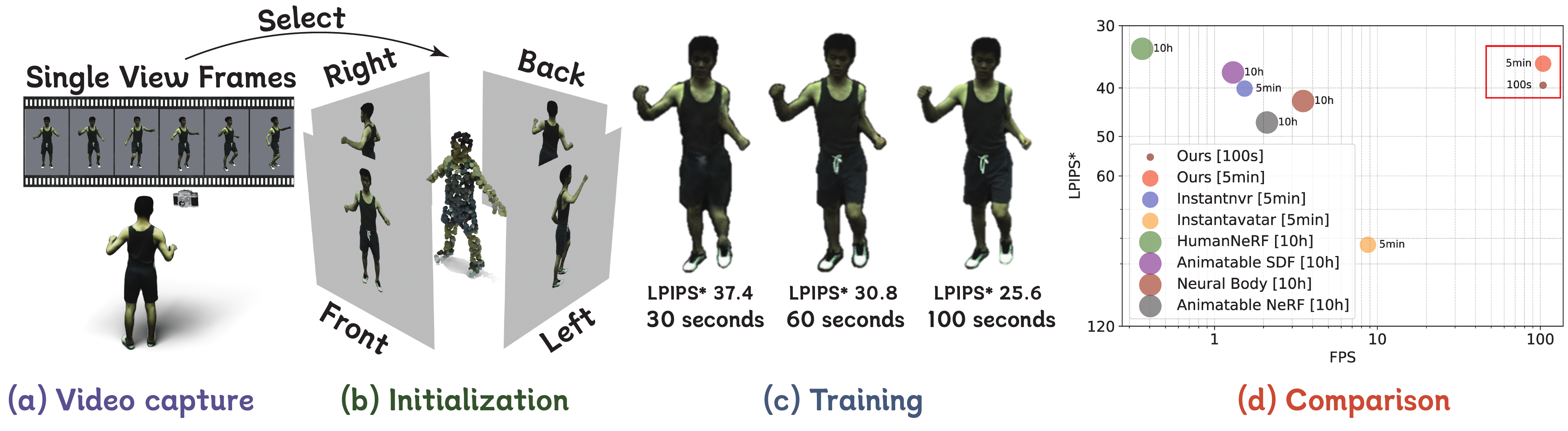}
    \captionsetup{type=figure}
    \vspace{-2.3em}
    \caption{\textbf{One VR common use-case.} (a) The user captures a short monocular video and proceeds to upload it. (b) Our model automatically selects four frames from the monocular video and after an initialization process, we can obtain an initial point cloud. (c) Our model can learn in minutes to get a dynamic human representation. (d) Our model achieves comparable or better visual quality while rendering much faster than previous works. LPIPS* = LPIPS $\times 10^3$. The area of each circle is proportional to the training time required, with larger areas representing longer training durations.}
    \vspace{-0.6em}
    \label{fig:teaser}
\end{center}

%% file: 00_abstract.tex
\begin{abstract}
Reconstructing the human body from single-view videos plays a pivotal role in the virtual reality domain. One prevalent application scenario necessitates the rapid reconstruction of high-fidelity 3D digital humans while simultaneously ensuring real-time rendering and interaction. Existing methods often struggle to fulfill both requirements. In this paper, we introduce Human101, a novel framework adept at producing high-fidelity dynamic 3D human reconstructions from 1-view videos by training 3D Gaussians in 100 seconds and rendering in 100+ FPS. Our method leverages the strengths of 3D Gaussian Splatting, which provides an explicit and efficient representation of 3D humans. Standing apart from prior NeRF-based pipelines, Human101 ingeniously applies a Human-centric Forward Gaussian Animation method to deform the parameters of 3D Gaussians, thereby enhancing rendering speed (i.e., rendering 1024-resolution images at an impressive 60+ FPS and rendering 512-resolution images at 100+ FPS). Experimental results indicate that our approach substantially eclipses current methods, clocking up to a 10 \( \times \) surge in frames per second and delivering comparable or superior rendering quality. Code and demos will be released at \href{https://github.com/longxiang-ai/Human101}{https://github.com/longxiang-ai/Human101}.
\end{abstract}

%% file: 01_intro.tex
\section{Introduction}
\label{sec:intro}
In the realm of virtual reality, a prevalent use case involves rapidly crafting custom virtual avatars and facilitating interactions with them. Within this context, two significant technical challenges emerge:
\textbf{(1)} How can we swiftly produce a digitized virtual avatar, preferably within a user's acceptable waiting time \textit{(e.g., within 3 minutes)}, using readily available equipment \textit{(e.g., a single-camera setup)}?
\textbf{(2)} How can we achieve real-time rendering to cater to the interactive demands of users?

While previous methods~\cite{peng2021neuralbody, peng2021animatablenerf, peng2022animatablesdf, Jiang_2023_CVPR_instantavatar, instant_nvr, zheng2023avatarrex, zhang2023explicifying} have made some progress, they still haven't fully met the requirements of the application scenario described earlier. The limitations of these methods can be summarized in two main points: \textbf{(1) Slow rendering speed in implicit methods.} Methods based on implicit neural network\cite{peng2021neuralbody, peng2021animatablenerf, peng2022animatablesdf, instant_nvr, Jiang_2023_CVPR_instantavatar} optimization using NeRF have slower rendering processes and challenges with inverse skinning deformation, preventing real-time rendering. \textbf{(2) Slow convergence speed in explicit methods.} Approaches such as those in \cite{zheng2023avatarrex, zhang2023explicifying}, capable of achieving real-time rendering, necessitate extensive data for training. This requirement results in slower optimization, thus hindering the rapid reconstruction of dynamic humans.

To address these challenges, a more practical and improved approach would fit these goals. 
\textbf{First}, to enhance rendering speed, we should choose a rasterization rendering pipeline, replacing the traditional volume rendering approach. 
\textbf{Second}, to speed up training, we should choose a better representation method that's easier to optimize, ideally reducing optimization time to just a few minutes. 
Recently, a novel method \cite{kerbl3Dgaussians} has employed 3D Gaussians to explicitly depict 3D scenes. With the integration of a differentiable tile rasterization method, it achieves superior visual quality and a much quicker rendering speed (over 100 FPS) compared to previous works~\cite{mueller2022instant, barron2022mipnerf360, yu_and_fridovichkeil2021plenoxels}. The emergence of this method makes realizing the described application scenario \textit{(i.e., achieving both fast reconstruction and real-time rendering)} a tangible possibility. 

Recognizing the advantages of \cite{kerbl3Dgaussians}, we introduce a novel framework for single-view human reconstruction. This framework not only accomplishes dynamic human reconstruction in less than one minute but also ensures real-time rendering capabilities. Merging the fast and straightforward methods of 3D GS with human body structures, we've created a new kind of \textbf{forward skinning process} for rendering. Different from the usual inverse skinning used by \cite{peng2021animatablenerf, peng2022animatablesdf, instant_nvr, Jiang_2023_CVPR_instantavatar} this forward skinning deformation method avoids searching for the corresponding canonical points of the target pose points but directly deform the canonical points into observation space. Cause \cite{kerbl3Dgaussians} utilizes 3D Gaussians rather than just points, we use a \textbf{Human-centric Forward Gaussian Animation} method to deform the positions, rotations, and scales of Gaussians, and modify spherical coefficients by rotating their directions. For faster convergence, we design a \textbf{Canonical Human Initialization} method to initialize the original Gaussians.

To validate the effectiveness of the proposed pipeline, we conduct extensive experiments on ZJU-MoCap Dataset~\cite{peng2021neuralbody} and the Monocap Dataset~\cite{instant_nvr}. Results show that Human101 could not only swiftly reconstruct a dynamic human, but also outperform incredible rendering speed together with better visual quality. With a single RTX 3090 GPU, our method can learn \textbf{in 100 seconds} to get comparable or better visual quality and maintain \textbf{100+ FPS rendering speed}, which makes it a tangible possibility for real-time interactive applications and immersive virtual reality experiences.

Our \textbf{contributions} can be summarised as follows:
\begin{itemize}
    \item We introduce an innovative approach to dynamically represent 3D human bodies by employing 3D Gaussian Splatting~\cite{kerbl3Dgaussians}, utilizing its efficient and explicit representation capabilities for detailed and accurate human modeling. We have proposed a \textbf{Canonical Human Initialization} method, which significantly enhances the model's convergence rate and improves visual detail representation.
    \item We propose a deformation methodology, composed of \textbf{Human-centric Forward Gaussian Animation} and \textbf{Human-centric Gaussian Refinement}, which is distinct from the prevailing time-consuming inverse skinning frameworks, making it possible to fast reconstruct and render a dynamic human in real time.
    \item We achieve a \textbf{$\sim $ 10.8 $ \times$} speed increase in rendering during inference (with \textbf{FPS 100+} for 512 $\times$ 512 resolution images) compared to previous neural human representations, while simultaneously ensuring comparable or superior rendering quality and higher image resolution.
\end{itemize}
\input{figs/tex/difference}

%% file: figs/tex/difference.tex
\begin{figure}
    \centering
    \includegraphics[width=\linewidth]{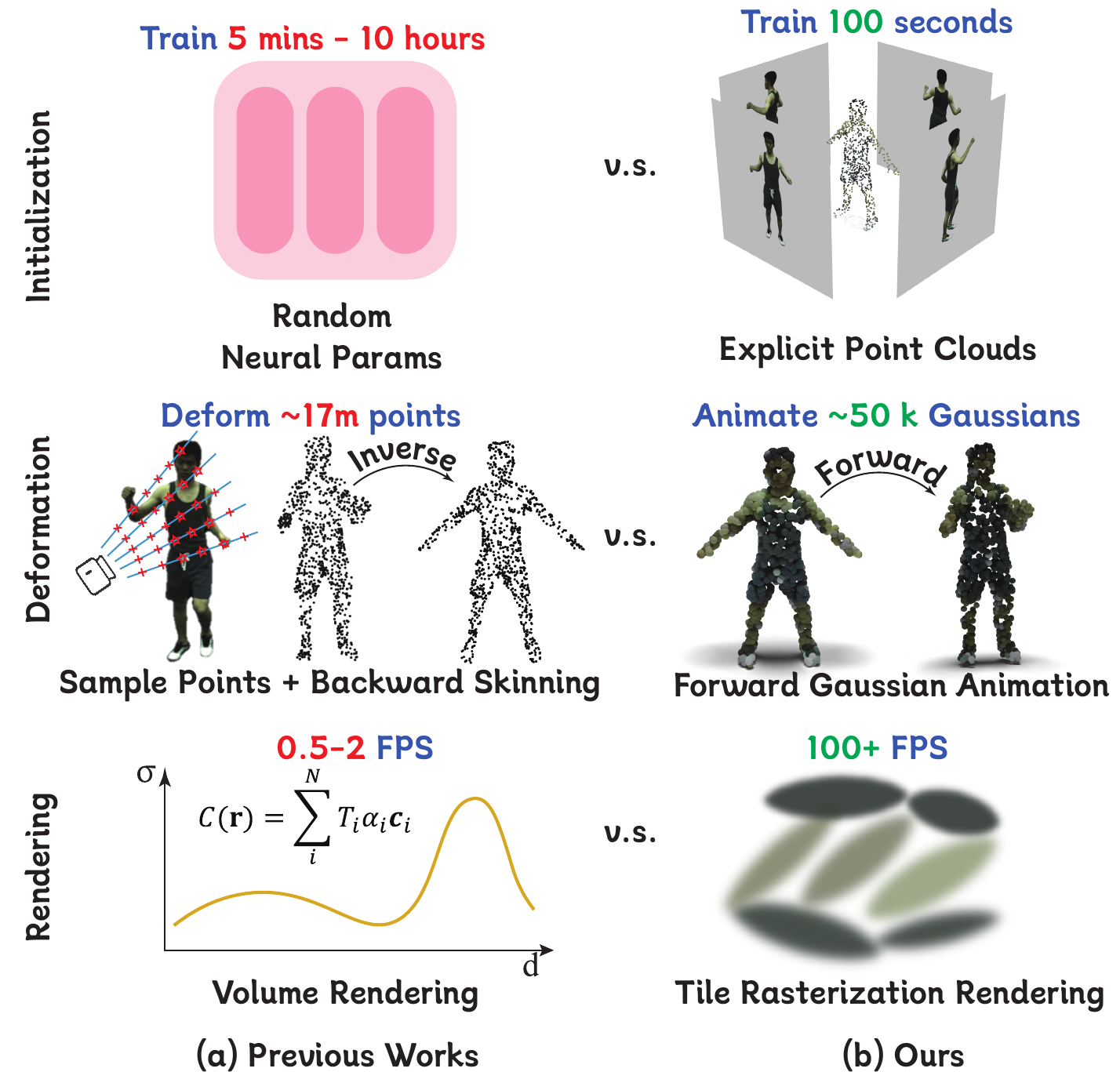}
    \captionsetup{type=figure}
    \vspace{-1.5em}
    \caption{\textbf{Method difference with previous works~\cite{instant_nvr, peng2021animatablenerf} 
 (\S~\ref{sec:method}).} \textbf{(1)} Initialization. Previous works rely on an implicit representation, initialized with random neural parameters. In contrast, our method employs explicit colored point clouds for initialization, which accelerates convergence. \textbf{(2)} Deformation. Earlier approaches require deforming a larger number of points from the target pose to the canonical pose, consuming more computational resources. \textbf{(3)} Rendering. Compared to the NeRF-based ray marching technique, our approach utilizing rasterization rendering achieves significantly faster rendering speeds.}
    \vspace{-1.5em}
    \label{fig:difference}
\end{figure}

%% file: 02_related.tex
\section{Related Work}
\label{sec:related}
\noindent\textbf{Human Reconstruction from Monocular Video.}
Reconstructing 3D humans from monocular videos is challenging, primarily due to complex human poses and the limited information from a single camera. Significant strides in this area have been made by \cite{kolotouros2019spin, ROMP1, ROMP2, ROMP3, jiang2022selfrecon, openpose1, openpose2, openpose3, openpose4, saito2019pifu, saito2020pifuhd, xiu2022icon, xiu2023econ, zhang2023global}. Approaches like \cite{peng2021neuralbody, peng2021animatablenerf, peng2022animatablesdf, weng_humannerf_2022_cvpr, jiang2022neuman} have excelled in high-quality reconstruction with precise pose and deformation adjustments. Yet, their lengthy convergence times, often exceeding 10 hours, limit practical utility.

Recent works like \cite{mueller2022instant, instant_nvr, Jiang_2023_CVPR_instantavatar} have accelerated convergence to about 5 minutes, but rendering speeds remain a bottleneck, with \cite{instant_nvr} achieving only about 1.5 FPS on an RTX 3090. 3D GS~\cite{kerbl3Dgaussians}, effective for static scenes with its explicit 3D Gaussian representation and fast GPU sorting, offers a potential solution with over 100 FPS rendering speeds. However, its application to dynamic human reconstruction is not straightforward. Our work harnesses the principles of 3D GS for monocular human reconstruction, targeting high rendering speeds to bridge the gap towards practical implementation in real-world applications.

\noindent\textbf{Human Deformation and Animation.}
The Skinned Multi-Person Linear model (SMPL)~\cite{SMPL:2015, SMPLIFY:2016, SMPL-X:2019} is a prevalent framework for representing human structure, simplifying pose changes with Linear Blend Skinning (LBS). Various generative articulation methods have been explored~\cite{hong2022avatarclip, zhang2022avatargen, bergman2022gnarf, ENARF, corona2021smplicit, chen2022gdna}, alongside backward skinning techniques~\cite{jeruzalski2020nilbs, LEAP:CVPR:21, tiwari21neuralgif, Saito:CVPR:2021} and forward skinning methods~\cite{chen2022gdna, lin2022fite, dong2022pina, jiang2022selfrecon, li2022tava, ARAH:ECCV:2022, chen2021snarf, Fast-SNARF_Chen2023PAMI}. For dynamic human reconstruction, studies like~\cite{peng2021animatablenerf, weng_humannerf_2022_cvpr, instant_nvr} use neural networks to enhance the deformation process, applying residuals to point coordinates or blending weights. \textbf{Unlike} these point-based approaches, our method employs 3D Gaussians~\cite{kerbl3Dgaussians} for spatial representation, accounting for position, rotation, and scale. Our human-centric forward skinning deformation approach successfully animates humans based on 3D Gaussians, effectively addressing challenges such as artifacts and jaggedness after deformation.

\noindent\textbf{Accelerating Neural Rendering.}
Since the introduction of NeRF by \cite{mildenhall2020nerf}, numerous studies have sought to accelerate neural scene rendering. Techniques utilize voxel grids \cite{yu2021plenoctrees,garbin2021fastnerf,bakingrealtime,yu_and_fridovichkeil2021plenoxels,reiser2021kilonerf,mueller2022instant,li2023nerfacc}, explicit surfaces \cite{chen2022mobilenerf,kulhanek2023tetra,patakin2023neural}, and point-based representations \cite{kopanas2021point,zhang2022differentiable,ruckert2022adop,rakhimov2022npbg++,lassner2021pulsar} or other methods to speed up rendering process. These methods effectively minimize the number of NeRF MLP evaluations required, thereby reducing computational costs. Focusing on human body, some approaches \cite{zheng2023avatarrex, Lombardi:2019:neuralvolume, sitzmann2019deepvoxels} utilize innovative processes like UV map prediction \cite{chen2023uv,kwon2023deliffas, zheng2023avatarrex, zhang2023explicifying} and voxel grids \cite{TiNeuVox}. However, these techniques predominantly suffer from lengthy training durations and are mostly restricted to static scenes, hindering downstream applications.

A significant breakthrough in this area is the development of 3D Gaussian Splatting (3D GS)~\cite{kerbl3Dgaussians}, utilizing anisotropic 3D Gaussians combined with spherical harmonics~\cite{muller2006spherical} to represent 3D scenes. This method effectively circumvents the slow ray marching operation, delivering high-fidelity and high-speed rendering. Nevertheless, its application has been primarily confined to static scenes. Our work pioneers the application of 3D GS to animatable human reconstruction. We extend the capabilities of 3D GS beyond static multi-view scenarios, addressing its limitations in dynamic monocular human movement reconstruction. 

%% file: 03_method.tex
\section{Method}
\label{sec:method}
\input{figs/tex/pipeline}
\noindent\textbf{Overview.}
In this work, our primary objective is to \textbf{swiftly} reconstruct dynamic human movements from \textbf{single-view videos}, simultaneously ensuring \textbf{real-time rendering} capabilities. Our approach builds upon the techniques introduced in~\cite{instant_nvr}, with the underlying assumption that the cameras are pre-calibrated and each image is accompanied by provided human poses and foreground human masks. \cref{fig:pipeline} shows the main training pipeline of our model. Within \cref{sec:preliminary}, we explore the foundational aspects of 3D Gaussian Splatting (3D GS)~\cite{kerbl3Dgaussians} and SMPL~\cite{SMPL:2015}. \cref{sec:canonicalinit} delves into the process of canonical human initialization, which is the base of the initial 3D Gaussians, speeding up our training process. 
\cref{sec:forwardanimation} presents our novel human-centric forward gaussian animation approach. This section details how we adapt and apply Gaussian models to accurately represent and animate human figures, focusing on achieving both high fidelity and efficiency in dynamic scenarios. Finally, \cref{sec:shsadaption} illustrates our refinements of Gaussians and spherical harmonics.

\subsection{Preliminary}
\label{sec:preliminary}
\noindent\textbf{3D Gaussian Splatting.}
Our framework employs 3D Gaussian Splatting~\cite{kerbl3Dgaussians} to parameterize dynamic 3D shapes for 2D image transformation. Differing from NeRF-based methods, we define 3D Gaussians with a full 3D covariance matrix $\Sigma$ centered at $\mu$ in world space. For 2D rendering, the projected covariance matrix $\Sigma'$ is calculated as:
\begin{equation}
    \Sigma' = J V \Sigma V^\top J^\top,
\end{equation}
where $J$ is the Jacobian of the affine approximation of the projective transformation, and $V$ is the world-to-camera matrix. To simplify learning, $\Sigma$ is decomposed into a quaternion $r$ for rotation and a 3D-vector $s$ for scaling, yielding rotation matrix $R$ and scaling matrix $S$. Hence, $\Sigma$ is expressed as:
\begin{equation}
    \Sigma = RSS^\top R^\top.
\end{equation}
\noindent\textbf{Adapting to dynamic scenes.}
The method proposed in \cite{kerbl3Dgaussians} excels in static scene representation from multi-view data. However, to extend this framework to dynamic human scenarios, we refine the 3D Gaussian model within the canonical space. We systematically optimize the Gaussians' defining attributes: \textit{position} \( x \), \textit{rotation} \( r \), \textit{scale} \( s \), and \textit{view direction} \( d \) of the radiance field, which is represented using spherical harmonics (SH). These refinements are carried out through an additional deformation field that enables the precise capture of the nuanced movements inherent to human dynamics. For clarity and conciseness, the 3D Gaussians in our framework are denoted as \(\mathcal{G}(x, r, s, d)\), succinctly encapsulating the parameters critical to modeling dynamic human forms.

\noindent\textbf{SMPL and LBS deformation.}
SMPL~\cite{SMPL:2015} is a widely used human skinning model, composed of 6890 vertices and 13776 triangular faces. Each of the SMPL vertices owns a 3D position $v_i$ and a corresponding weight vector $w_i$. The deformation of the SMPL mesh $\mathcal{M}$ is performed using the Linear Blend Skinning (LBS) technique, which deforms the mesh based on a set of pose parameters $\boldsymbol{\theta}$ and shape parameters $\boldsymbol{\beta}$. Specifically, Given $\boldsymbol{\theta}$ and $\boldsymbol{\beta}$, the LBS technique can transform vertex $v_{can}$ from canonical space to observation space $v_{ob}$ as follows:
\begin{equation}
    v_{ob} = \sum_{i=1}^{N}w_{i}G_i(\boldsymbol{\theta}, \boldsymbol{\beta}) v_{can},
\end{equation}
where $N$ is the joint number, $w_i$ is the blend weight of $v$, and $G_i(\boldsymbol{\theta}, \boldsymbol{\beta})$ is the transformation matrix of joint $i$.

\subsection{Canonical Human Initialization}
\label{sec:canonicalinit}
\input{figs/tex/cano_init}
\noindent\textbf{Previous Point Cloud Extraction.}
The initialization of the canonical Gaussian space is critically dependent on the point cloud data quality, with variations in initial point clouds substantially influencing the model's convergence speed and the refinement of outcomes. While prior studies~\cite{kerbl3Dgaussians} have adeptly derived initial point clouds from multi-view data leveraging COLMAP techniques~\cite{10.1145/1141911.1141964}, these methods excel predominantly in static scenes. \textbf{However}, such approaches do not perform well with a single viewpoint and are incapable of estimating point clouds for dynamic data.

\noindent\textbf{Human-centric Point Cloud Parts Extraction.}
Consequently, taking human structure into account, we consider the adoption of monocular reconstruction methods~\cite{xiu2022icon, xiu2023econ, zhang2023global} to acquire the initial point cloud. The prior state-of-the-art~\cite{xiu2023econ} is capable of estimating a mesh from a single image input and projecting the image color onto the mesh. However, this approach only attaches the color of one image to the predicted mesh, resulting in a mesh that is only partially colored. Our goal is to select as few images as possible from the input sequence to achieve the best initial results. Our \textbf{Automatic Selection} strategy involves extracting multiple sets of images from the sequence, each set containing four images with the human subject's angles as close to 90 degrees apart as possible. We then choose one set of images where the poses most closely resemble the ``A'' pose (shown in the \cref{fig:pipeline} (a) ), which facilitates mesh deformation into the canonical pose in the subsequent steps. These images are labeled according to their orientation: \textit{front} \( \mathbf{F} \), \textit{back} \( \mathbf{B} \), \textit{left} \( \textbf{L} \), and \textit{right} \( \textbf{R} \). Using~\cite{xiu2023econ}, we generate four meshes \(\mathcal{M}^F(\mathcal{V}^F), \mathcal{M}^B( \mathcal{V}^B), \mathcal{M}^L(\mathcal{V}^L), \mathcal{M}^R(\mathcal{V}^R)\), each representing different postures and having color only on one side.

\noindent\textbf{Canonical Point Cloud Fusion.}
These meshes are deformed to a canonical pose using inverse Linear Blend Skinning~\cite{SMPL:2015}.
\begin{equation}
    \mathcal{V}_{\text{can}}^k = \left( \sum_{i=1}^{N} w_i^k G_i^k \right) ^{-1} \mathcal{V}^k ,\text{ for } k \in \{\mathbf{F}, \mathbf{B}, \mathbf{L}, \mathbf{R}\}
\end{equation}
\begin{equation}
    \mathcal{P}_{\text{can}}(\mathcal{V}_{\text{can}}) = \text{Fuse}(\mathcal{M}_{\text{can}}^F, \mathcal{M}_{\text{can}}^B, \mathcal{M}_{\text{can}}^L, \mathcal{M}_{\text{can}}^R)
\end{equation} 
As a result, each canonical pose point cloud part is colored on one side only. We then fuse the four canonical point clouds to form an initial point cloud in canonical space, comprising approximately 50,000 points. 
Following~\cite{kerbl3Dgaussians}, we convert the initial point cloud into canonical Gaussians. Additionally, our model can also initialize using the bare SMPL's canonical pose mesh (6890 vertices with white colors). While this approach requires a slightly longer convergence time, it still ensures a visually appealing result of comparable quality. 



\subsection{Human-centric Gaussian Forward Animation}
\label{sec:forwardanimation}
\noindent \textbf{Advantages of Gaussian Approach.}
Traditional NeRF-based methods~\cite{peng2021animatablenerf, peng2022animatablesdf, instant_nvr, Jiang_2023_CVPR_instantavatar}, involving inverse LBS for deformation and ray sampling in observation space, face efficiency issues at high resolutions (e.g., \(512 \times 512 \times 64 \approx 16, 777k\)), as depicted in \cref{fig:difference}. These methods struggle with real-time rendering due to the vast number of sampling points and the slow inverse LBS deformation process.

Our approach deviates from these conventional methods by optimizing explicit Gaussians in canonical space and animating them into observation space. This method results in more efficient rendering, especially suitable for dynamic scenes.

\noindent \textbf{Human-centric Gaussian Translation.}
Given the \textit{original Gaussian position} \( {x} \), the \textit{transformation matrix} of the \( i \)-th bone \( G_i(\boldsymbol{\theta}, \boldsymbol{\beta}) \), and the \textit{blend weight} of the \( i \)-th bone \( w_i \). The \textit{deformed Gaussian position} \( x' \) is represented as:
\begin{equation}
    {x'} = \sum_{i=1}^{N} w_i G_i(\boldsymbol{\theta}, \boldsymbol{\beta}) {x},
\end{equation}where \(G_i(\boldsymbol{\theta}, \boldsymbol{\beta})\) is computed by SMPL vertex which is the nearest to the \( i \)-th Gaussian.

\noindent \textbf{Human-centric Gaussian Rotation.}
Gaussians inherently display \textbf{anisotropic} characteristics, meaning they exhibit different properties in various directions. Therefore, accurate rotation adjustments are crucial when transitioning between canonical and observation poses, ensuring the model's adaptability across new observation viewpoints. 

Unlike positions, rotations cannot be directly derived using Linear Blend Skinning (LBS). 
To define the rotation of Gaussians, we anchor each Gaussian to the SMPL mesh by identifying the closest triangular facet based on Euclidean distance. The position of a triangular facet's centroid, denoted as \( f_j^p \), is calculated as the mean of its vertices \( f_j^1, f_j^2, \) and \( f_j^3 \):
\begin{equation}
    f_j^p = \frac{f_j^1 + f_j^2 + f_j^3}{3}.
\end{equation}
For the \( i \)-th Gaussian, the nearest triangle facet is determined by finding the minimum distance to \( x_i \):
\begin{equation}
    j^* = \underset{j}{\mathrm{argmin}} \, \| x_i - f_j^p \|.
\end{equation}
We then adopt the rotation \( R_{f_{j^*}} \) of this facet as the Gaussian's rotation transformation matrix, which we called Triangular Face Rotation:
\begin{equation}
    R_i = R_{f_{j^*}} = e_{\text{can}_{j^*}} e_{\text{ob}_{j^*}}^\top,
\end{equation}
\begin{equation}
    r'_i = \text{Rot}(R_i, r_i),
\end{equation}
where \(r'_i\) is rotated Gaussian rotation, \(e_{\text{can}_j^*}\) and \(e_{\text{ob}_j^*}\) represent the orthonormal bases of the facet in the canonical and observed poses, respectively. These bases are computed from the edge vectors' normalized cross products. To enhance computational efficiency during training, we precompute \( R_{f} \) leveraging the known distribution of SMPL poses.

\noindent\textbf{Rotation of Spherical Harmonics.}
To ensure that the rotation of spherical harmonics aligns with the human body's posture, we rotate these functions during rendering. As a part of Gaussians' attributes, spherical harmonics should rotate together with Gaussians to precisely represent the view-dependent colors. Given the rotation of \( i \)-th Gaussian \( R_i \) , we can easily get rotated spherical harmonics by applying \( R_i \) into its direction \( d_i\), that is:
\begin{equation}
    d'_i = \text{SH\_Rot}(R_i, d_i) = R_i^{\top} d_i
\end{equation}

\subsection{Human-centric Gaussian Refinement}
\label{sec:shsadaption}
\input{figs/tex/trick}
\noindent\textbf{Adaptive Gaussian Refinement.}
To capture the uniqueness of each frame, we use the \textit{frame index} \( t \) in \textit{Positional Encoding (PE)} to get \textit{per-frame feature} \(\boldsymbol{\gamma} (t)\). The \textit{Gaussian positions}, represented as \( {x} \), are fed through a \textit{Multilayer Perceptron (MLP)} to calculate position, rotation, and scale residuals. These are then used to adjust the Gaussian parameters for each frame, ensuring accuracy and consistency.The adaptive adjustments are represented as follows:
\begin{equation}
     \Delta {x}, \Delta {r}, \Delta {s} = F_{\boldsymbol{\Theta}} \left(\boldsymbol{\gamma}(x), \boldsymbol{\gamma}(t), \boldsymbol{\theta}, \boldsymbol{\beta} \right) ,
\end{equation}
where $\boldsymbol{\theta}, \boldsymbol{\beta}$ are the parameters of SMPL \textit{pose} and \textit{shape}. This results in deformed parameters: \textit{position} \( {x''} = {x'} + \Delta {x} \), \textit{rotation} \( {r''} = \text{Rot}(\Delta {r} , {r'}) \), and \textit{scale} \( {s''} = {s} + \Delta {s} \).

\noindent\textbf{Spherical Harmonics Refinement.}
Similar to the refinement process of Gaussians, given \(\Delta r_i \) of the \( i \)-th Gaussian, we can easily get refined Spherical harmonics by this formula:
\begin{equation}
    d''_i = \text{SH\_Rot}(\Delta r_i, d'_i) = \text{quat\_to\_rotmat}(\Delta r_i)^{\top} d'_i
\end{equation}
Thus, we get the refined Gaussians \(\mathcal{G}(x'', r'', s'', d'')\), and send them to the fast rasterization rendering module of ~\cite{kerbl3Dgaussians} to get high-fidelity results. 
\subsection{Data Augmentation Technique}
During our experimental process, we noted that spherical harmonics coefficients tended to overfit when limited to inputs from a single, fixed camera perspective. This resulted in pronounced color biases, as depicted in \cref{fig:ablation}. To mitigate this issue, we adopted a data augmentation approach that enhances the diversity of camera perspectives.
With SMPL global orientation \textit{rotation matrix} \(R_s\), \textit{global translation matrix} \(T_s\) and \textit{camera pose} \(R_c\), \(T_c\). We can get the following equation, describing how SMPL coordinates \(x_s\) are transformed into camera coordinates \( x_c\) :
\begin{equation}
    x_c = R_c(R_sx_s+T_s)+T_c .
    \label{eq:notrick}
\end{equation}
If we assume that the SMPL coordinates align with the world coordinates, (i.e. \(R_s' = E\), \(T_s' = O\)), we can get:
\begin{equation}
    x_c = R_c'x_s+T_c'.
    \label{eq:trick}
\end{equation}
With \cref{eq:notrick} and \cref{eq:trick}, we can easily get:
\begin{equation}
    R_c' = R_cR_s, T_c' = R_cT_s+T_c.
\end{equation}
This method effectively simulates the camera encircling the subject, as shown in \cref{fig:trick}, leading to a more varied orientation in the spherical harmonics. It successfully diversifies the orientations represented by the spherical harmonics, preventing overfitting and the associated color distortions.

\input{tables/all_comparison}
\subsection{Training}
\label{sec:training}
\noindent\textbf{Setting.}
Our model takes single-view camera parameters, SMPL parameters, frame indices, and images as inputs. During training, frames are randomly sampled from the video. The predicted image, denoted as \( \hat{I} \), is constrained using the L1 loss and S3IM~\cite{xie2023s3im} loss. In contrast to the method in~\cite{instant_nvr}, we chose not to use the VGG loss due to its high computational demands, which could slow down our training process. For now, we have also decided against incorporating regularization terms. The loss function is formulated as:
\begin{equation}
\mathcal{L}_{rgb} = \lambda_1  ||\hat{I} - I_{\text{gt}}|| + \lambda_2 \text{S3IM}(\hat{I}, I_{\text{gt}})
\end{equation}
where \( \lambda_1 \) and \( \lambda_2 \) are the weighting factors for L1 and S3IM loss respectively, and \( I_{\text{gt}} \) is the ground truth image.

\subsection{Implementation Details}
\label{sec:implementation}
Following~\cite{kerbl3Dgaussians, instant_nvr}, we employ the Adam optimizer~\cite{kingma2017adam}, setting distinct learning rates for different parameters. Our model is trained on an RTX 3090 GPU, and it seamlessly reaches a level of performance comparable to previous methods in just about 100 seconds. Following this, the model requires roughly 10,000 iterations, culminating in convergence within about 5 minutes. Notably, while our model excels at single-view reconstruction, it further enhances accuracy and results when applied to multi-view reconstructions, all within the same 5-minute timeframe. For an in-depth comparison of multi-view reconstructions, we direct interested readers to the supplementary materials. Additionally, a comprehensive overview of our network structure and hyper-parameters can also be found in the supplementary section.

%% file: figs/tex/pipeline.tex
\begin{figure*}[h!tp]
    \centering
    \includegraphics[width=\linewidth]{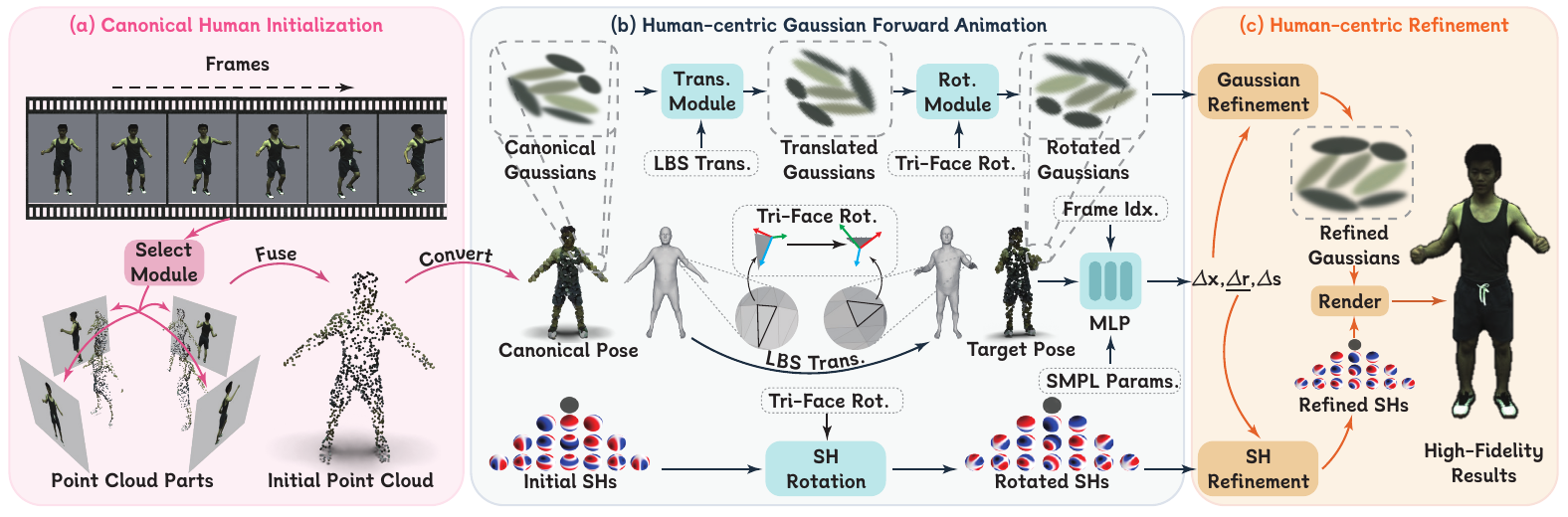}
    \vspace{-1.5em}
    \caption{
    \textbf{Overview of Human101 (\S~\ref{sec:method}).} \textbf{(a) Canonical Human Initialization (\S~\ref{sec:canonicalinit})}. We use an offline model~\cite{xiu2023econ} to extract 4 point cloud parts from 4 selected frames, and fuse them into a canonical point cloud, which then is converted into canonical Gaussians. \textbf{(b) Human-centric Gaussian Forward Animation (\S~\ref{sec:forwardanimation})}. We deform canonical 3D Gaussians into the target pose by modifying Gaussian positions \(x\), rotations \(r\), and scales \(s\)). And we rotate the spherical coefficients with triangle face rotation. \textbf{(c) Human-centric Gaussian Refinement (\S~\ref{sec:shsadaption}).} We refine positions \(x\), rotations \(r\) and scales \(s\) of Gaussians and refine the view direction \(d\) of spherical harmonics. 
    }
    \vspace{-1.5em}
    \label{fig:pipeline}
\end{figure*}

%% file: figs/tex/trick.tex
\begin{figure}
    \centering
    \includegraphics[width=\linewidth]{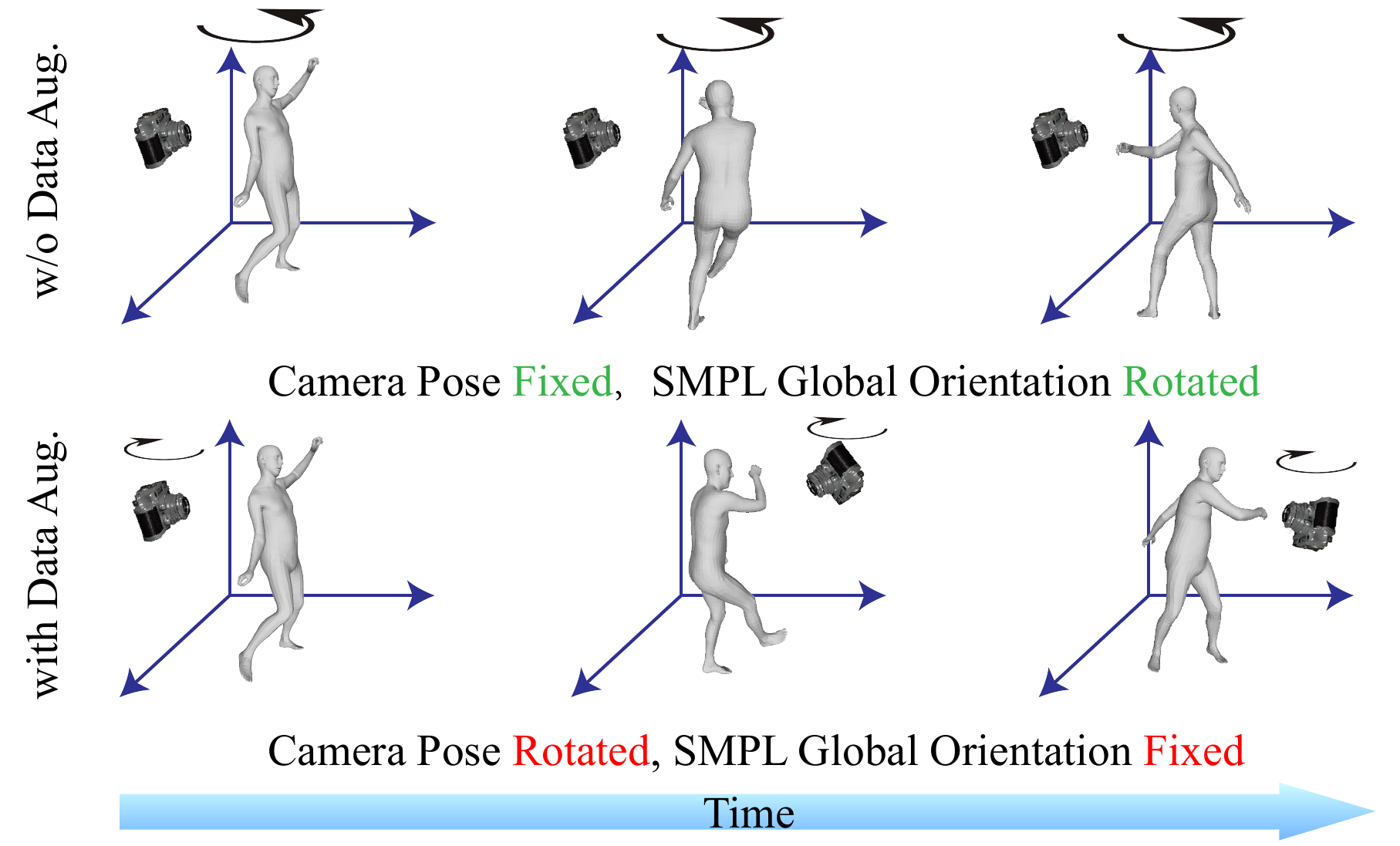}
    \captionsetup{type=figure}
    \vspace{-1.5em}
    \caption{\textbf{Data Augmentation Technique (\S~\ref{sec:shsadaption}).} Through this method, we simulate the camera rotating around the human to make up multiple camera poses.}
    \label{fig:trick}
    \vspace{-1.3em}
\end{figure}

%% file: tables/all_comparison.tex
\begin{table*}
\centering
\scriptsize
\begin{tabular}{rrrl|cccc|cccc}
    \Xhline{2\arrayrulewidth}
    \rowcolor[HTML]{C0C0C0} 
    \multicolumn{4}{l|}{} &
    \multicolumn{4}{c|}{ZJU-MoCap~\cite{peng2021neuralbody}} &
    \multicolumn{4}{c}{MonoCap~\cite{instant_nvr}} \\ 
    
    \rowcolor[HTML]{C0C0C0} 
    {Method} &
    Publication & 
    Res. &
    Train &
    PSNR$\uparrow$ &
    SSIM$\uparrow$ &
    LPIPS*$\downarrow$ &
    FPS$\uparrow$ &
    PSNR$\uparrow$ &
    SSIM$\uparrow$ &
    LPIPS*$\downarrow$ &
    FPS$\uparrow$ 
    \\ \Xhline{2\arrayrulewidth}
    
    \multicolumn{12}{c}{\textit{Static scene reconstruction method}} 
    \\ \hline
    
    3D GS + SMPL Init.  ~\cite{kerbl3Dgaussians}
    &
        SIGGRAPH 
    23 &
    512 &
    $\sim $ 5min &
    26.57 &
    0.935 &
    71.70 &
    156   &
    28.47 &
    0.972 &
    29.57 &
    156   \\
    3D GS + SMPL Init.  ~\cite{kerbl3Dgaussians}
    &
        SIGGRAPH 
    23 &
    1024 &
    $\sim $ 5min &

    26.53 &
    0.944 &
    58.23 &
    51.3  &
    27.47 &
    0.970 &
    34.37 &
    51.3  
    \\
    
    \hline
    \multicolumn{12}{c}{\textit{Time-consuming human reconstruction method}} \\
    \hline
    NeuralBody ~\cite{peng2021neuralbody}
    &
        CVPR
    21 &
    512 &
    $\sim $ 10h &

    29.03 &
    0.964 &
    42.47 &
    3.5 &
    32.36 &
    0.986 &
    16.70 &
    3.5 \\
    
    AnimNeRF ~\cite{peng2021animatablenerf}
    &
        ICCV
    21 &
    512 &
    $\sim $ 10h &

    29.77 &
    0.965 &
    46.89 &
    2.1 &
    31.07 &
    0.985 &
    19.47 &
    2.1 \\
    AnimSDF  ~\cite{peng2022animatablesdf}
    &
        Arxiv
    22 &
    512 &
    $\sim $ 10h &

    30.38 &
    0.975 &
    37.23 &
    1.3 &
    32.48 &
    0.988 &
    13.18 &
    1.3 \\
    HumanNeRF  ~\cite{weng_humannerf_2022_cvpr}
    &
            CVPR
    22 &
    512 &

    $\sim $ 10h &

    30.66 &
    0.969 &
    33.38 &
    0.36 &
    32.68 &
    0.987 &
    15.52 &
    0.36 \\ \hline
    \multicolumn{12}{c}{\textit{Time-efficient human reconstruction method}} \\
    \hline
    InstantAvatar ~\cite{Jiang_2023_CVPR_instantavatar}
    &
        CVPR
    23 &
    512 &
    $\sim $ 5min &

    29.21 &
    0.936 &
    82.42 &
    \secondcolor8.75    &
    32.18 &
    0.977 &
    24.98 &
    \secondcolor8.75
    \\
    InstantNvr ~\cite{instant_nvr}
    &
        CVPR
    23 &
    512 &
    $\sim $ 5min &

    30.87 &
    \bestcolor 0.971 &
    40.11 &
    1.53  &
    32.61 &
    \bestcolor 0.988 &
    16.68 &
    1.53
    \\ \hline
    \textbf{Ours}
    &
     
    &
    512 &
    $\sim $ \textbf{100s} &
     \secondcolor 31.29
    & 0.964
    & \secondcolor 39.50
    & \bestcolor 104
    & \bestcolor 33.20
    & \secondcolor 0.983
    & \secondcolor 16.55
    & \bestcolor 104
    \\
    \textbf{Ours}
    &  &
    512 &
    $\sim $ 5min &
    \bestcolor 31.79 &
    \secondcolor 0.965 &
    \bestcolor 35.75 &
    \bestcolor 104 &
    \secondcolor32.63 &
    0.982 &
    \bestcolor16.51 & \bestcolor 104
    \\
    \hline
    \hline
    InstantAvatar~\cite{Jiang_2023_CVPR_instantavatar} 
    &
        CVPR
    23 &
    1024 &
    $\sim $ 5min &

    27.79 &
    0.912 &
    97.33 &
    \secondcolor3.83 &
    32.10 &
    \secondcolor0.978 &
    24.95 &
    \secondcolor3.83  \\
    InstantNvr~\cite{instant_nvr} 
    &
        CVPR
    23 &
    1024 &
    $\sim $ 5min &

    30.89 &
    \bestcolor 0.974 &
    41.70 &
    0.54 &
    \textit{OOM} &
    \textit{OOM} &
    \textit{OOM} &
    \textit{OOM}
    \\
    \hline
    \textbf{Ours}
    &
        
    &
    1024 & 
    $\sim $ \textbf{100s} &

    \bestcolor 31.00 &
    \secondcolor 0.968 &
    \secondcolor 40.71 &
    \bestcolor 68 &
    \bestcolor 32.20  &
    \bestcolor 0.983  &
    \secondcolor 18.31  &
    \bestcolor 68
    \\ 
    \textbf{Ours}
    &
        
    &
    1024 &
    $\sim $ 5min &

    \secondcolor 30.93 &
    0.967 &
    \bestcolor 39.87 &
    \bestcolor 68 &
    \secondcolor 32.13 &
    \bestcolor 0.983 &
    \bestcolor 17.01 &
    \bestcolor 68
    \\
    \Xhline{2\arrayrulewidth}
\end{tabular}
\vspace{-0.5em}
    \caption{\textbf{Comparison with SOTA (\S~\ref{sec:comparewithsota}).} We compare Human101 with several baseline methods. 
    (1) Static scene reconstruction method: 3D GS~\cite{kerbl3Dgaussians}. 
    (2) Time-consuming human reconstruction methods: HumanNeRF~\cite{weng_humannerf_2022_cvpr}, AnimSDF~\cite{peng2022animatablesdf}, NeuralBody~\cite{peng2021neuralbody} and AnimNeRF~\cite{peng2021animatablenerf}. 
    (3) Time-efficient human reconstruction methods: InstantNvr~\cite{instant_nvr}, InstantAvatar~\cite{Jiang_2023_CVPR_instantavatar}. 
    LPIPS* = LPIPS $\times 10^3$, and ``\textit{OOM}'' means out of GPU memory when training. 
    For the FPS metric, we evaluate by calculating the inference time provided by the official pre-trained models. We have marked out \colorbox{yzybest}{best} and \colorbox{yzysecond}{second best} metrics of time-efficient human reconstruction methods.}
    \label{tab:all_comparison}
    \vspace{-1.3em}
\end{table*}

%% file: 04_experiments.tex
\vspace{-1.0em}
\section{Experiments}
\input{figs/tex/compare}
\label{sec:experiments}
\subsection{Datasets}
\noindent \textbf{ZJU-MoCap Dataset.} ZJU-Mocap~\cite{peng2021neuralbody} is a prominent benchmark in human modeling from videos, supplying foreground human masks and SMPL parameters. Similar to~\cite{instant_nvr}, our experiments engage 6 human subjects (377, 386, 387, 392, 393, 394) from the dataset. Training utilizes one camera, while the remaining cameras are designated for evaluation. Each subject contributes 100 frames for training.

\noindent \textbf{MonoCap Dataset}. The MonoCap Dataset combines four multi-view videos from the DeepCap~\cite{deepcap} and DynaCap~\cite{habermann2021} datasets, collected by~\cite{peng2021animatablenerf}. This dataset provides essential details like camera parameters, SMPL parameters and human masks. We choose the same 4 subjects as~\cite{instant_nvr} for better comparison. Further details about all the sequences in our study can be found in the supplementary material.

\subsection{Comparison with the state-of-the-art methods}
\label{sec:comparewithsota}
\textbf{Baselines.} We compare our method with some previous human reconstruction methods~\cite{peng2021neuralbody, peng2021animatablenerf, peng2022animatablesdf, instant_nvr, Jiang_2023_CVPR_instantavatar} and baseline method 3D GS~\cite{kerbl3Dgaussians}. The methods for human reconstruction can be categorized into 3 groups:
\begin{itemize}
    \item \textbf{Static scene reconstruction method.} 3D GS~\cite{kerbl3Dgaussians} is the backbone of our model. While its original point cloud initialization method failed to extract a point cloud from a single fixed-view camera, we use the canonical SMPL vertices together with white colors as its initial point cloud.
    \item \textbf{Time-consuming human reconstruction methods.} NeuralBody~\cite{peng2021neuralbody} utilizes structured SMPL data together with per-frame latent codes to optimize neural human radiance fields. AnimatableNeRF(AnimNeRF)~\cite{peng2021animatablenerf} and AnimatableSDF(AnimSDF)~\cite{peng2022animatablesdf} use SMPL deformation and pose-dependent neural blend weight field to model dynamic humans. HumanNeRF~\cite{weng_humannerf_2022_cvpr} further optimizes volumetric human representations, and improves detail quality of rendered image. However, due to the slow optimization of MLPs, these methods usually converge very slow. For instance,~\cite{weng_humannerf_2022_cvpr} takes about 72 hours on 4 RTX 2080Ti GPUs to totally converge.
    \item \textbf{Time-efficient human reconstruction methods.} Utilizing~\cite{mueller2022instant}'s voxel grid representation, Instantnvr~\cite{instant_nvr} manage to shorten the convergence time into 5 minutes. InstantAvatar~\cite{Jiang_2023_CVPR_instantavatar} combines ~\cite{mueller2022instant} with a rapid deformation method~\cite{Fast-SNARF_Chen2023PAMI}, achieving a fast reconstruction in minutes. 
\end{itemize}
\noindent\textbf{Metrics.} We choose PSNR, SSIM, LPIPS~\cite{zhang2018perceptual} as visual quality evaluation metrics, and frame per second (FPS) as rendering speed evaluation metrics. For better comparison, we show LPIPS* = LPIPS \( \times 10 ^ 3\) instead. \cref{tab:all_comparison} shows our results compared with others. Here we list only the average metric values of all selected characters on a dataset due to the size limit while more detailed qualitative and quantitative comparisons are in the supplementary material. 

\input{figs/tex/lpips_compare}
\input{figs/tex/ablation}

\noindent\textbf{Discussion on quantitative results.}
\cref{tab:all_comparison} presents a comprehensive quantitative comparison between our method and other prominent techniques like InstantNvr~\cite{instant_nvr}, InstantAvatar~\cite{Jiang_2023_CVPR_instantavatar}, 3D GS~\cite{kerbl3Dgaussians}, HumanNeRF~\cite{weng_humannerf_2022_cvpr}, AnimSDF~\cite{peng2022animatablesdf}, NeuralBody~\cite{peng2021neuralbody}, and AnimNeRF~\cite{peng2021animatablenerf}. Remarkably, our approach achieves optimization within approximately 100 seconds, yielding results that are comparable with or surpass ~\cite{instant_nvr, Jiang_2023_CVPR_instantavatar, kerbl3Dgaussians} in terms of PSNR and LPIPS. For 5 minutes' results, we achieved the highest PSNR and LPIPS scores among rapid reconstruction methods. While our SSIM scores are quite high, they do not reach the state-of-the-art level. This is partly attributed to the characteristics of spherical harmonics in our model. Spherical harmonics, by their nature, are somewhat limited in capturing high-frequency color information. Moreover, for inference speed, our model is \textbf{67 times faster} than InstantNvr~\cite{instant_nvr} and \textbf{11 times faster} than InstantAvatar~\cite{Jiang_2023_CVPR_instantavatar} in 512 \( \times \) 512 resolution.

\noindent\textbf{Discussion on qualitative results.}
\cref{fig:compare} showcases a comparison of our model with time-efficient reconstruction works ~\cite{instant_nvr, Jiang_2023_CVPR_instantavatar, kerbl3Dgaussians}. Our method stands out by providing the most detailed representation and minimal artifacts, as highlighted in \cref{fig:compare} with red boxes around key details. In contrast, ~\cite{instant_nvr}'s backside results exhibit unnatural colors on the body's front due to light penetration and loss of details like missing logos on trousers. InstantAvatar~\cite{Jiang_2023_CVPR_instantavatar} generates noticeable scattered artifacts around the body. Meanwhile, 3D GS~\cite{kerbl3Dgaussians}, lacking a deformation module for dynamic scenes, results in severe limb truncations and facial distortions. Furthermore, in terms of rendering speed, except for~\cite{kerbl3Dgaussians}, our inference speed surpasses all compared methods.

\subsection{Ablation Study}
\label{sec:ablation}
\input{tables/ablation_components}
\input{tables/ablation_init}
Our ablation study results on the ZJU-MoCap~\cite{peng2021neuralbody} 377 sequence are displayed in \cref{tab:ablation_components}, \cref{tab:ablationinit}, and \cref{fig:ablation} (a). Due to the space limitations, more experiment results can be found in the supplementary material.

\noindent\textbf{Human-centric Gaussian rigid deformation.} 
Omitting it,\textit{ (i.e. without SMPL-based rigid translation and rotation, w/o SMPL Deform.)}, leads to noticeable deficiencies in body parts and a loss of detail, as shown in \cref{fig:ablation} (a). This experiment highlights the limitations of simple SMPL deformation in fully capturing complex human motions. The results underscore the necessity of integrating more sophisticated deformation techniques to accurately model human movements.

\noindent \textbf{Spherical rotation.}
The results, illustrated in \cref{fig:ablation} (a), highlight that omitting spherical rotation leads to ``spiky'' artifacts on the human body, compromising visual quality and causing abnormal lighting effects in the rendered images.

\noindent \textbf{Data augmentation.}
Our data augmentation technique is crucial, as evidenced by overfitting of spherical harmonics and noticeable skin artifacts in novel view synthesis when it's absent, as depicted in \cref{fig:ablation} (a).

\noindent \textbf{Adaptive Gaussian Refinement.}
The lack of the Adaptive Gaussian Refinement module results in the model's inability to capture subtle human deformations (like in fingers), leading to visible artifacts (shown in \cref{fig:ablation} (a)).

\noindent\textbf{Initialization method.}
Our comprehensive study demonstrates the effectiveness of our novel canonical human initialization method. As shown in \cref{fig:ablation} (b), this method significantly enhances the quality of human body reconstruction. 

%% file: figs/tex/compare.tex
\begin{figure*}[!h!t!p]
    \centering
    \includegraphics[width=\linewidth]{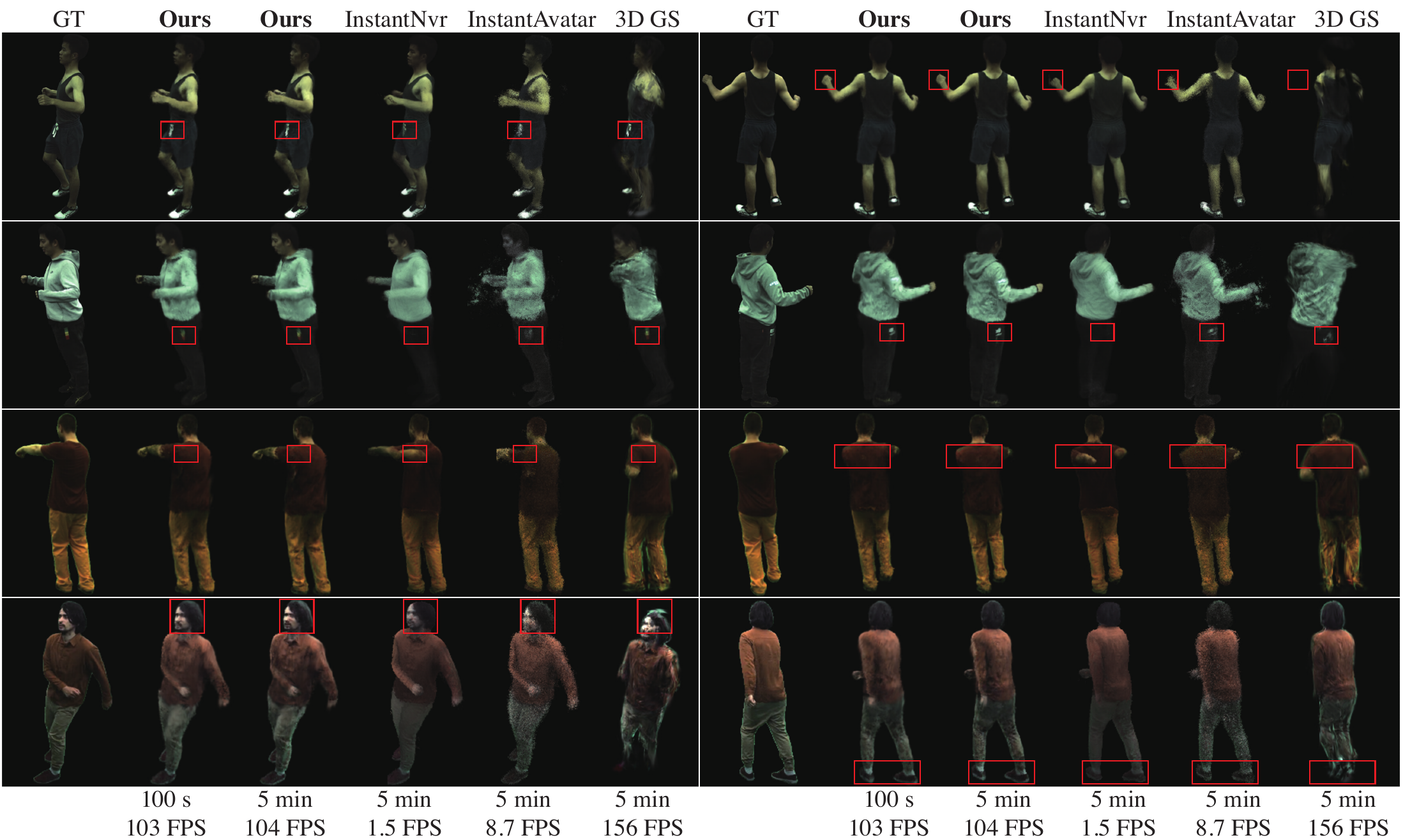}
    \vspace{-1.3em}
    \caption{
    \textbf{Compare with the state-of-the-art works (\S~\ref{sec:comparewithsota}).} For a fair comparison, we show the results of 512 \( \times\) 512 resolutions. Our model delivers results with superior visual quality and richer details, achieving a 67 $\times$ increase in FPS compared with the state-of-the-art time-efficient methods. Please \faSearch ~zoom in for a more detailed observation. 
    }
    \label{fig:compare}
    \vspace{-1.5em}
\end{figure*}

%% file: figs/tex/lpips_compare.tex

%% file: figs/tex/ablation.tex
\begin{figure*}[!htp]
    \centering
    \includegraphics[width=\linewidth]{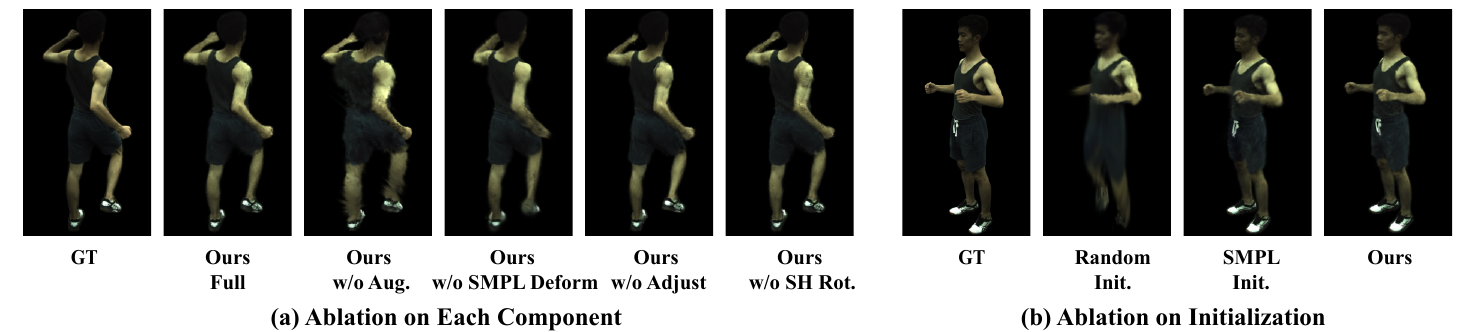}
    \vspace{-1.5em}
    \caption{\textbf{Ablation studies on the Sequence 377 of ZJU-MoCap dataset~\cite{peng2021neuralbody} (\S~\ref{sec:ablation}).} 
    (a) Removing our proposed components leads to and blurry appearance and artifacts.
    (b) Initialization plays a pivotal role in the geometry quality.
    }
    \label{fig:ablation}
    \vspace{-1.5em}
\end{figure*}

%% file: tables/ablation_components.tex

\begin{table}
    \centering
    \scriptsize
    \begin{tabular}{l|ccc}
    \Xhline{2\arrayrulewidth}
    \rowcolor[HTML]{C0C0C0} 
    Method              & PSNR $\uparrow$                          & SSIM $\uparrow$                         & LPIPS* $\downarrow$                       \\ \hline
    w/o SMPL Deformation    & 28.70                         & \secondcolor 0.964                         & 41.61                         \\
    w/o Augmentation       & 29.00                         & 0.961                         & 41.30                         \\
    w/o Frame Embedding & \secondcolor 32.26                         & \bestcolor 0.977                         & \secondcolor 22.51                         \\
    w/o Gaussian Adjustment      & \bestcolor 32.55 & \bestcolor 0.977                         & 23.20                         \\
    \textbf{Ours Full}                & 32.18                         & \bestcolor0.977 & \bestcolor21.32
    \\ 
    \Xhline{2\arrayrulewidth}
    \end{tabular}
    \vspace{-1.2em}
    \caption{\textbf{Ablation study on component efficacy (\S~\ref{sec:ablation}).} This table demonstrates the impact of individual components in our method. By selectively disabling each part, we validate their effectiveness. The experiments were conducted on Sequence 377 of the ZJU-MoCap dataset~\cite{peng2021neuralbody}.}
    \label{tab:ablation_components}
\end{table}

%% file: tables/ablation_init.tex
\begin{table}
    \centering
    \scriptsize
\begin{tabular}{l|ccc}
\Xhline{2\arrayrulewidth}
\rowcolor[HTML]{C0C0C0} 
Method      & PSNR $\uparrow$  & SSIM $\uparrow$ & LPIPS* $\downarrow $\\ \hline
Random Init. & 26.98 & 0.951 & 62.35  \\
SMPL Init.   & \secondcolor 31.84 & \secondcolor 0.974 & \secondcolor 28.60  \\
\textbf{Ours} & \bestcolor32.18 & \bestcolor0.977 & \bestcolor21.32 \\ 
\Xhline{2\arrayrulewidth}
\end{tabular}
\vspace{-1.2em}
    \caption{\textbf{Ablation on initialization method (\S~\ref{sec:ablation}).} We use Sequence 377 on ZJU-MoCap\cite{peng2021neuralbody} to test the effectiveness of our canonical human initialization process.}
    \label{tab:ablationinit}
    \vspace{-1.5em}
\end{table}

%% file: 10_conclusion.tex
\section{Conclusion}
\label{sec:conclusion}
In this paper, we introduced Human101, a novel framework for single-view human reconstruction using 3D Gaussian Splatting (3D GS)~\cite{kerbl3Dgaussians}. Our method efficiently reconstructs high-fidelity dynamic human models within just 100 seconds using a fixed-perspective camera. The integration of a novel Canonical Human Initialization, Human-centric Gaussian Forward Animation, and Human-centric Gaussian Refinement, coupled with 3D GS's explicit representation, significantly improve the rendering speed. Moreover, this enhancement in speed does not sacrifice visual quality. Experiments demonstrate that Human101 outperforms up to 67 times in FPS compared with the state-of-the-art methods and maintain comparable or better visual quality.
Human101 sets a new standard in human reconstruction from single-view videos. This breakthrough lays the groundwork for further advances and applications in immersive technologies.

%% file: 12_appendix.tex
\maketitlesupplementary

\section{Overview}
Overview of the Supplementary Material:

\begin{itemize}
    \item Implementation Details \S~\ref{sec:sup_implementation}:
    \begin{itemize}
        \item Conventions in Symbolic Operations. \S~\ref{sec:sup_conventions}
        \item Dataset. \S~\ref{sec:sup_dataset}
        \item Baseline Implementation Details. \S~\ref{sec:sup_base_implementation}
        \item Hyperparameters. \S~\ref{sec:sup_hyperparam}
        \item Network Structure. \S~\ref{sec:sup_netstructure}
        \item Canonical Human Initialization. \S~\ref{sec:sup_initialization}
        \item Details of Triangular Face Rotation Matrices. \S~\ref{sec:sup_tri_face_rot}
    \end{itemize}
    
    \item Additional Quantitative Results \S~\ref{sec:sup_quantitative}:
    \begin{itemize}
        \item Novel View Results. \S~\ref{sec:sup_novelview_quan}
        \item Multi-view Results. \S~\ref{sec:sub_multi-view_quan}
    \end{itemize}

    \item Additional Qualitative Results \S~\ref{sec:sup_qualitative}:
    \begin{itemize}
        \item Depth Visualization Results. \S~\ref{sec:sup_depth}
        \item Novel Pose Results. \S~\ref{sec:sup_novelpose}
    \end{itemize}

    \item More Experiments \S~\ref{sec:sup_experiments}:
    \begin{itemize}
        \item Memory Efficiency Comparison. \S~\ref{sec:sup_efficiency_compare}
        \item Ablation Study. \S~\ref{sec:sup_ablation}
        \item Failure Cases.  \S~\ref{sec:sup_failure_cases}
    \end{itemize}

    \item Downstream Applications \S~\ref{sec:sup_application}:
    \begin{itemize}
        \item Composite Scene Rendering Results. \S~\ref{sec:sup_composite}
    \end{itemize}
    
    \item More Discussions \S~\ref{sec:sup_discussion}:
    \begin{itemize}
        \item Data Preprocessing Technique. \S~\ref{sec:sup_datapreprocess}
        \item Limitations. \S~\ref{sec:sup_limitations}
        \item Ethics Considerations. \S~\ref{sec:sup_ethicsconsiderations}
    \end{itemize}
\end{itemize}

\section{Implementation Details}
\label{sec:sup_implementation}

\subsection{Conventions in Symbolic Operations}
\label{sec:sup_conventions}
In our work, the rotation operations involve various types of rotational quantities (such as rotation matrices and quaternions). For simplicity, we represent these rotation operations in the format of ``multiplication" in the main text. Here, we detail this representation more concretely:

For Gaussian rotation \( r_i \), when optimizing \( r_i \), it is considered a quaternion. While rotating it by the triangular face rotation matrix \( R_i \), we first convert \( R_i \) into a unit quaternion and express this process using quaternion multiplication. Thus, this operation is denoted as:
\begin{equation}
    r'_i = \text{quat\_multi} \left(\text{rotmat\_to\_quat}(R_i), r_i  \right)
\end{equation}

When applying rotation in the refinement module, the predicted \( \Delta r_i \) is a quaternion. Therefore, this rotation is expanded as:
\begin{equation}
    r''_i = \text{quat\_multi}(\Delta r_i, r'_i)
\end{equation}

For spherical harmonics, modifying spherical coefficients directly is not efficient. A more effective approach is to inversely rotate the view direction \( d_i \). Specifically, we first calculate the direction from the camera center \( P_c \) to the final Gaussian position \( x''_i \) as the view direction. Then, we apply the inverse rotation transformation to view directions as the input for SH evaluation. Specifically, we have:
\begin{equation}
    d_i = x''_i - P_c
\end{equation}
\begin{equation}
    d'_i = \text{SH\_Rot}(R_i, d_i)
\end{equation}
\begin{equation}
    d''_i = \text{SH\_Rot}(\text{quat\_to\_rotmat}(\Delta r_i), d'_i)
\end{equation}

The function \(\text{SH\_Rot}\) takes a rotation matrix and a view direction as input, returning a rotated view direction:
\begin{equation}
    \text{SH\_Rot}(R, d) = R^{-1} d = R^\top d
\end{equation}

\subsection{Dataset}
\label{sec:sup_dataset}
\input{tables/dataset}
\noindent \textbf{ZJU-MoCap.} For ZJU-MoCap Dataset~\cite{peng2021neuralbody}, we choose 6 subjects (377, 386, 387, 392, 393, 394) for evaluation.  Because other subjects tend to not appear on the full side in a single fixed view. And following~\cite{instant_nvr}, we use camera 04 for training and other views for testing. Due to the low quality of the images in camera 03 for Subject 377 and Subject 392, we filter out these two views.

\noindent \textbf{MonoCap.} MonoCap is re-collected by~\cite{peng2022animatablesdf}, with Lan \& Marc 1024 \( \times \) 1024 resolution, selected from DeepCap dataset~\cite{deepcap} and olek \& vlad 1295 \( \times \) 940 resolution selected from~\cite{habermann2021}. For better comparison, we show the FPS results of Lan and Marc. The DeepCap dataset~\cite{deepcap} and DynaCap dataset~\cite{habermann2021} are
only granted for non-commercial academic purposes. They prohibit the redistribution of that data. The users should also sign a license. More frame-selecting details are illustrated in~\cref{tab:dataset_settings}.

\subsection{Baseline Implementation Details}
\label{sec:sup_base_implementation}
For Neural Body~\cite{peng2021neuralbody}, Animatable NeRF~\cite{peng2021animatablenerf}, and AnimatableSDF~\cite{peng2022animatablesdf}, we utilized the results released in~\cite{instant_nvr}. We also tested their rendering speeds by inferring with their pre-trained models on the same device using a single RTX 3090 GPU.

The work presented in~\cite{Jiang_2023_CVPR_instantavatar} did not have an implementation for the ZJU-MoCap and MonoCap Datasets due to their slightly varied SMPL definition. Consequently, we adjusted the deformer in~\cite{Jiang_2023_CVPR_instantavatar} to match the SMPL vertices. It's important to note that~\cite{Jiang_2023_CVPR_instantavatar} is designed specifically for monocular datasets, and it refines the SMPL parameters before metric evaluation. For a fair comparison, we adhered to the same SMPL and camera parameters provided by the ZJU-MoCap Dataset and MonoCap Dataset, as with other baseline methods~\cite{instant_nvr, peng2021neuralbody, peng2021animatablenerf, peng2022animatablesdf}, and chose not to refine the SMPL parameters before evaluation.

Regarding the 3D Gaussian Splatting~\cite{kerbl3Dgaussians}, COLMAP could not determine valid camera parameters due to the input of monocular fixed-view video frames. As a solution, we opted to use the SMPL vertices from the initial frame as the input point cloud positions and designated the point cloud colors as white. Given that~\cite{kerbl3Dgaussians} is primarily a static multi-view 3D reconstruction method, achieving convergence in our setup proved challenging. Hence, we present the outcomes at 30k iterations, consistent with its original settings.

\subsection{Hyperparameters}
\label{sec:sup_hyperparam}
We experimentally fine-tuned our model employing a set of hyperparameters tailored for optimal performance. 
Regarding the spherical harmonics, we employed third-degree spherical harmonics for their balance of computational efficiency and representational fidelity. Uniquely, we increment the degree of spherical harmonics every 500 iterations, culminating at a maximum degree of three. For the learnable MLP component, we set the learning rate at \(2 \times 10^{-3}\). During the optimization of Gaussians, we implemented an opacity reset at every 1500 iterations to refine transparency values.

\subsection{Network Structure}
\label{sec:sup_netstructure}
\input{figs/tex/network_structure}
To compensate for the rigid position and rotation using the learnable MLP, we employ straightforward linear layers, featuring a total of 5 layers with \( n_{\text{hidden\_dim}} = 64 \). ReLU serves as the activation function between these layers, while no activation function is applied to the output. For SMPL parameters, we use a simple linear layer to compress its feature dimension. We use Positional Encoding with a frequency of 10. \cref{fig:network_structure} demonstrates the structure of the linear networks.

\subsection{Canonical Human Initialization}
\label{sec:sup_initialization}
In the initialization process, we use an algorithmic approach instead of manually selecting four photos. Our objective is to select four images where the person's angles on each are approximately 90 degrees apart. Additionally, it is preferable that the person's pose in these images closely resembles the canonical pose. This ensures minimal accuracy loss when deforming the point cloud estimated by econ into the canonical pose. To achieve this, we undertake the following steps:

\noindent \textbf{1. Identify suitable image pairs.} We traverse the dataset's frames and for each frame index in frame index T, we maintain a set \( C_i \), \(C_i\) records all frame indices whose angle \(\delta\) with frame index $i$ is between 80-100 degrees. The formula is as follows:
\begin{equation}
    C_i = \{ j \mid 80 \leq \delta_{ij} \leq 100, \ \forall j \neq i \ \text{and} \ j > i \}, \ \forall i \in T
\end{equation}
The angle \(\delta_{ij}\) between frames \(i\) and \(j\) is derived by calculating the difference in angles of the global rotation matrices \( R_{\text{global}} \) from the SMPL parameters of the two frames. The formula is as follows:
\begin{equation}
    R_{\text{diff\ ij}} = R_{\text{global\ i}}^{-1} \cdot R_{\text{global\ j}}
\end{equation}
\begin{equation}
    \delta_{ij} = \text{as\_euler}(R_{\text{diff\ ij}})
\end{equation}
\noindent \textbf{2. Select a suitable group of frames.} The second part involves identifying a set of four images that meet the criteria, executed through a four-level nested loop. Initially, frame \(i\) is selected, followed by choosing \(j\) from the set \(C_i\). Subsequently, \(k\) is selected from \(j\)'s set \(C_j\), and \(l\) from \(C_k\). For each selected group of frames \( (i, j, k, l) \), the algorithm first checks if the angular difference between every two frames exceeds 80 degrees. Then, it computes the distance between the pose's joint positions in these images and the joint positions of the canonical pose. Finally, the group of frames with the smallest distance is selected. The process is shown in \cref{alg:frame_selection}.
\begin{algorithm}
\caption{Frame Selection (\S~\ref{sec:sup_initialization})}
\begin{algorithmic}
\label{alg:frame_selection}
\STATE {\textbf{Data:} Sets $T, C$}
\STATE {\textbf{Result:} Best frame indices $I_{best}$ and minimum distance $d_{min}$}
\STATE {$d_{min} \gets \infty$}
\STATE {$I_{best} \gets \emptyset$}
\FOR{$i \in T$}
    \FOR{$j \in C_i$}
        \FOR{$k \in C_j$}
            \FOR{$l \in C_k$}
                \IF{$\delta_{ik} > 80^\circ$ \textbf{and} $\delta_{il} > 80^\circ$ \textbf{and} $\delta_{jl} > 80^\circ$}
                    \STATE {$d \gets \text{distance}(\text{pose}(i, j, k, l), \text{canonical pose})$}
                    \IF{$d < d_{min}$}
                        \STATE {$d_{min} \gets d$}
                        \STATE {$I_{best} \gets \{i, j, k, l\}$}
                    \ENDIF
                \ENDIF
            \ENDFOR
        \ENDFOR
    \ENDFOR
\ENDFOR
\end{algorithmic}
\end{algorithm}

\subsection{Details of Triangular Face Rotation Matrices}
\label{sec:sup_tri_face_rot}
The process of computing rotation matrices involves two main steps: first, determining the orthonormal basis vectors (\(e_{\text{can}}\) and \(e_{\text{ob}}\)) that describe the orientation of each triangular facet of the SMPL model in the canonical and target poses, respectively; second, constructing the rotation matrix from these basis vectors.

For each triangular facet \(f\) constituted by vertices \(A\), \(B\), and \(C\), and edges \(AB\), \(AC\), and \(BC\), we define the first unit direction vector as:
\begin{equation}
    \overrightarrow{a} = \frac{\overrightarrow{AB}}{\lVert \overrightarrow{AB} \rVert}  
\end{equation}
Then, we use the normal of the triangular plane as the second unit direction vector:
\begin{equation}
    \overrightarrow{b} = \frac{\overrightarrow{AB} \times \overrightarrow{AC}}{\lVert \overrightarrow{AB} \times \overrightarrow{AC} \rVert}  
\end{equation}
Subsequently, the third direction vector is derived from the cross-product of the first two unit vectors:
\begin{equation}
    \overrightarrow{c\vphantom{b}} = \overrightarrow{a\vphantom{b}} \times \overrightarrow{b\vphantom{b}}
\end{equation}

Combining these vectors, we obtain the orthonormal basis for the triangular facet:
\begin{equation}
    e = ( \overrightarrow{a\vphantom{b}}, \overrightarrow{b}, \overrightarrow{c\vphantom{b}} )
\end{equation}

Having acquired the orthonormal bases in both canonical and observation spaces, the triangular face rotation matrix is computed as:
\begin{equation}
    R_f = e_{\text{can}} e_{\text{ob}}^\top
\end{equation}

%

\section{Additional Quantitative Results}
\label{sec:sup_quantitative}

\subsection{Novel View Results}
\label{sec:sup_novelview_quan}
For the novel view setting, \cref{tab:sup_novel_view_512} and \cref{tab:sup_novel_view_1024} show our results separately for resolution 512 $\times$ 512 and 1024 $\times$ 1024.

\subsection{Multi-view Results}
\label{sec:sub_multi-view_quan}
While our model was not specifically designed for multi-view training data, we have conducted tests on the ZJU-MoCap Dataset to assess its performance in such scenarios. The results, as depicted in \cref{fig:sup_multi_view}, demonstrate the model's capability to handle multi-view inputs.

\input{figs/tex/sup_depth_visiualization}
\input{figs/tex/sup_multiview}
\input{figs/tex/sup_novel_pose}

\input{tables/novel_view_1024_sepa}
\input{tables/novel_view_512_sepa}
\input{tables/novel_pose_1024_sepa}
\input{tables/novel_pose_512_sepa}
\input{tables/novel_view_1024_mutli_view}

\section{Additional Qualitative Results}
\label{sec:sup_qualitative}
\subsection{Depth Visualization}
\label{sec:sup_depth}
As shown in \cref{fig:sup_depth}, our method, with its explicit representation, achieves a superior depth representation. This illustrates the advantages of our approach in terms of geometric accuracy.

\subsection{Novel Pose Results}
\label{sec:sup_novelpose}
The results of our model trained on Subject 377 for unseen poses are shown in \cref{fig:sup_novel_pose}. Compared to the outcomes from InstantNvr, our results are less prone to artifacts and unnatural limb distortions. Simultaneously, our color reproduction is closer to the ground truth, with more preserved details in image brightness.

\section{Additional Experiments}
\label{sec:sup_experiments}
\subsection{Memory Efficiency Comparison}
\label{sec:sup_efficiency_compare}
\input{tables/efficiency_comparison}
To assess the efficiency of our model, we compared its resource consumption during the inference process with recent works in the field. In our comparison, we focused on three key metrics: training time GPU memory consumption (``Train Memory''), inference GPU memory consumption (``Infer Memory''), and disk space required for storing the model checkpoints (``Model Size''). In our work, the model size is computed by the sum of point cloud size and MLP checkpoint size. 

As illustrated in \cref{tab:mem_efficiency_comparison}, Human101 demonstrates notable memory efficiency compared to prior methods~\cite{instant_nvr, Jiang_2023_CVPR_instantavatar}. During training, our model employs a strategy aligned with downstream applications, opting for direct run-time querying of the neural network for rendering. This decision not only conserves space but also facilitates real-time rendering capabilities, as opposed to pre-storing query results which would increase storage requirements and impede real-time performance. 



\subsection{Ablation Study}
\label{sec:sup_ablation}
\noindent \textbf{Sparse Input Frames.}
Our model consistently delivers impressive results even with fewer input video frames. For Subject 377, as detailed in \cref{tab:ablation_sparse_frame}, we showcase our performance metrics for varying frame counts, specifically at 250, 100, 50, and 25 frames.
\input{tables/ablation_sparse_frame}

\noindent \textbf{Positional Encoding.}
In our experiments, we explored different positional encoding strategies for Gaussian positions, specifically comparing Instant-ngp~\cite{mueller2022instant}'s grid encoding against the traditional sine and cosine positional encoding. While grid encoding can experimentally accelerate the fitting process on the training frames, it also tends to make the model more susceptible to overfitting. Consequently, as demonstrated in~\cref{tab:ablation_PE}, this results in suboptimal performance on novel view test frames.
\input{tables/ablation_PE}

\noindent \textbf{Degree of Spherical Harmonics.}
\input{tables/ablation_shs}
We have also performed ablation experiments to determine the optimal degree of spherical harmonics for our reconstruction task. As indicated by \cref{tab:sup_ablation_shs}, increasing the degree of spherical harmonics leads to improved reconstruction quality. However, higher degrees bring a greater computational load. Consequently, we have chosen to adopt third-degree spherical harmonics for fitting in our final model, balancing accuracy with computational efficiency.

\noindent \textbf{Converge Speed on different Initialization.} 
\input{figs/tex/sup_converge_speed}
\input{figs/tex/gaussian_count}
\cref{fig:sup_converge_speed} demonstrates that the choice of initialization method significantly impacts the model's convergence speed. Furthermore, \cref{fig:sup_gaussian_count} shows different initialization strategies result in varying numbers of Gaussians at convergence. Generally, for the same scene, a larger number of Gaussians at convergence corresponds to richer reconstructed details. Since querying the MLP network is the more time-consuming factor during the inference phase, an increase in the number of Gaussians does not substantially affect the rendering FPS.

\subsection{Failure Cases}
\label{sec:sup_failure_cases}
Our model adeptly processes both monocular and multi-view video inputs, achieving high-fidelity reconstructions from sparse view inputs within a brief training duration. However, it is important to acknowledge the model's limitations. In instances where the input video fails to provide precise masks — for example, during intense movement where flowing hair carries unmasked background elements — this can result in visual artifacts, as depicted in \cref{fig:failure_case}.

\input{figs/tex/failure_case}
\input{figs/tex/composite}
\section{Application}
\label{sec:sup_application}

\subsection{Composite Scene Rendering}
\label{sec:sup_composite}

Rendering a human figure against a plain color background alone is not ideal for further downstream applications. Thanks to the explicit representation capability of 3D Gaussian Splatting (3D GS)~\cite{kerbl3Dgaussians}, we can effortlessly segregate dynamic human figures from static scenes by explicitly splicing the Gaussians. This splicing process is natural and allows for the easy separation of static backgrounds and dynamic human elements.

As demonstrated in \cref{fig:composite}, this functionality facilitates downstream applications. In the example, the background and the human subject are trained separately and then composited during the rendering process. See supplementary videos for better results.







\section{More Discussions}
\label{sec:sup_discussion}



\subsection{Discussions on Data Preprocessing Technique}
\label{sec:sup_datapreprocess}
Given that our task operates within a single-camera setting, we empirically observed during our experiments that, within fixed-view monocular videos, spherical harmonic coefficients tend to overfit to a singular direction. This leads to subpar generalization for free-view videos, resulting in numerous artifacts. To address this, we employed a data augmentation strategy that mimics a multi-camera environment. With access to the SMPL parameters detailing the global rotation of the human subject, it's intuitive to keep the human orientation static while allowing the camera to orbit around the figure. This mimics a nearly equivalent process. Using this technique, we simulate varying camera viewpoints to render the dynamic human across different frames, markedly boosting the generalizability of the spherical harmonic functions.

However, this trick isn't devoid of limitations. In real-world scenarios, due to the diffuse reflection of light, we often perceive varying colors for the same object from different viewpoints. Our strategy overlooks this variance, providing an approximation that might not always align perfectly with real-world lighting conditions.

\subsection{Limitations}
\label{sec:sup_limitations}
While Human101 marks a significant advancement in dynamic human reconstruction, it is not without its limitations:
\begin{itemize}
    \item Dependency on SMPL parameter accuracy. Human101 is significantly affected by the accuracy of SMPL parameter estimation. Inaccurate parameters can introduce substantial noise, complicating the reconstruction process.
    \item Requirement for complete body visibility in training data. Our model achieves the best results when training data includes all body parts relevant to the task. Partial visibility, where some body parts are not fully captured, may lead to artifacts in the reconstructed model.
\end{itemize}
Addressing these limitations could involve integrating more comprehensive human body priors, providing a pathway for future enhancements to our framework.

\subsection{Ethics Considerations}
\label{sec:sup_ethicsconsiderations}
Ethical considerations, particularly around privacy, consent, and data security, are critical in the development and application of Human101. Ensuring informed consent for all participants and transparent communication about the project's capabilities and limitations is essential to respect privacy and avoid misrepresentation. Secure handling and storage of sensitive human data are paramount to prevent unauthorized access and misuse. Additionally, acknowledging the potential for misuse of this advanced technology, we emphasize the need for ethical guidelines to govern its responsible use. Our commitment is to uphold high ethical standards in all aspects of Human101, safeguarding the respectful and secure use of human data.
\subsection{Broader Impact}
\label{sec:sup_broaderimpact}
The development of Human101 has significant implications across various domains. Its ability to rapidly reconstruct high-quality, realistic human figures from single-view videos holds immense potential in fields such as virtual reality, animation, and telepresence. This technology can enhance user experiences in gaming, film production, and virtual meetings, offering more immersive and interactive environments. However, its potential misuse in creating deepfakes or violating privacy cannot be ignored. It's crucial to balance innovation with responsible use, ensuring that Human101 serves to benefit society while minimizing negative impacts. Ongoing dialogue and regulation are necessary to navigate the ethical challenges posed by such advanced technology. Overall, Human101 stands to make a substantial impact in advancing digital human modeling while prompting necessary discussions on technology's ethical use.

%% file: tables/dataset.tex
\begin{table*}[]
\centering
\footnotesize
\begin{tabular}{r|c|c|c|c|c|c}
\Xhline{2\arrayrulewidth}
\rowcolor[HTML]{C0C0C0} 
Dataset                   & subject & Training view & Testing view index                           & Start Frame & End Frame & Frame Interval \\
\Xhline{2\arrayrulewidth}
                            & 386, 387, 393, 394 & 4  & Remaining                                & 0     & 500   & 5 \\
\multirow{-2}{*}{ZJU-MoCap~\cite{peng2021neuralbody}} & 377, 392           & 4  & Remaining except 3                       & 0     & 500   & 5 \\
                            & Lan                & 0  & Remaining                                & 620   & 1120  & 5 \\
                            & Marc               & 0  & Remaining                                & 35000 & 35500 & 5 \\
                            & Olek               & 44 & 0, 5, 10, 15, 20, 25, 30, 35, 40, 45, 49 & 12300 & 12800 & 5 \\
\multirow{-4}{*}{MonoCap~\cite{peng2021animatablenerf}} & Vlad    & 66            & 0, 10, 20, 30, 40, 50, 60, 70, 80, 90, 100 & 15275       & 15775     & 5  \\
\Xhline{2\arrayrulewidth}
\end{tabular}

\caption{\textbf{Dataset settings} (\S~\ref{sec:sup_dataset}).}
\label{tab:dataset_settings}
\end{table*}

%% file: figs/tex/network_structure.tex
\begin{figure*}
    \centering
    \includegraphics[width=0.7\linewidth]{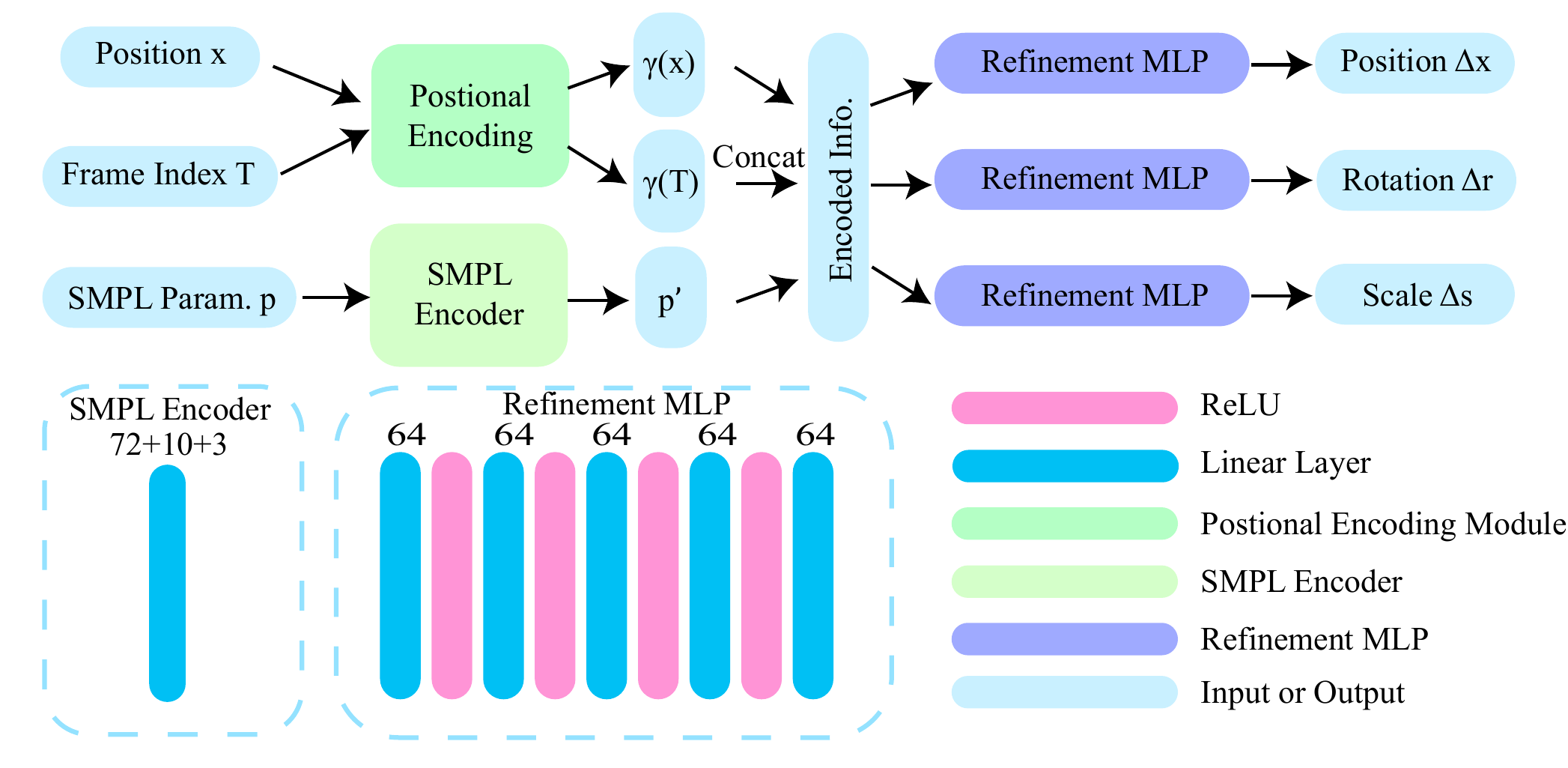}
    \captionsetup{type=figure}
    \caption{\textbf{Network structure (\S~\ref{sec:sup_netstructure}).} This diagram presents the network architecture for refining the attributes of Gaussians. The input position \( x \), frame index \( T \), and SMPL parameters \( p \) are first processed through positional encoding and an SMPL encoder, respectively. The encoded information \( \gamma(x) \), \( \gamma(T) \), and \( p' \) are then concatenated and passed through a series of refinement MLPs to produce adjustments in position \( \Delta x \), rotation \( \Delta r \), and scale \( \Delta s \). Each refinement MLP is composed of linear layers and employs ReLU activation functions.}
    \label{fig:network_structure}
\end{figure*}

%% file: figs/tex/sup_depth_visiualization.tex
\begin{figure*}
    \centering
    \includegraphics[width=\linewidth]{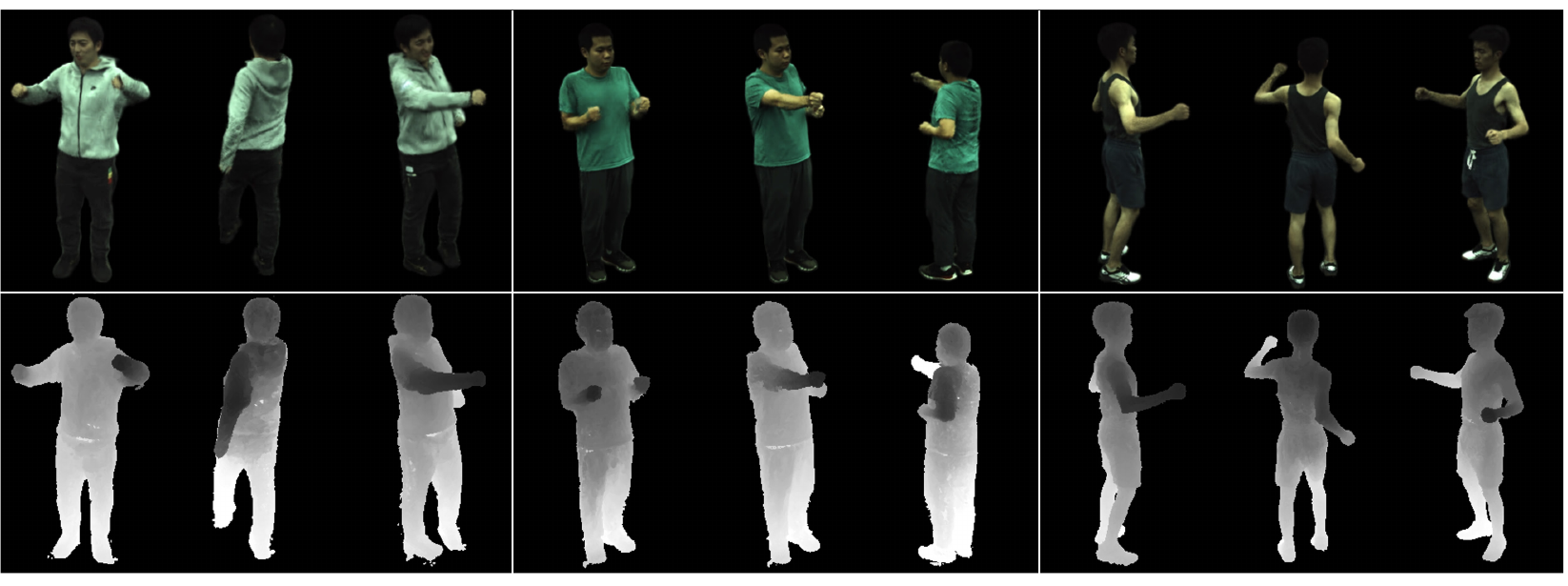}
    \captionsetup{type=figure}
    \vspace{-1.5em}
    \caption{\textbf{Depth visualization results (\S~\ref{sec:sup_depth}).}}
    \label{fig:sup_depth}
\end{figure*}

%% file: figs/tex/sup_multiview.tex
\begin{figure*}
    \centering
    \includegraphics[width=\linewidth]{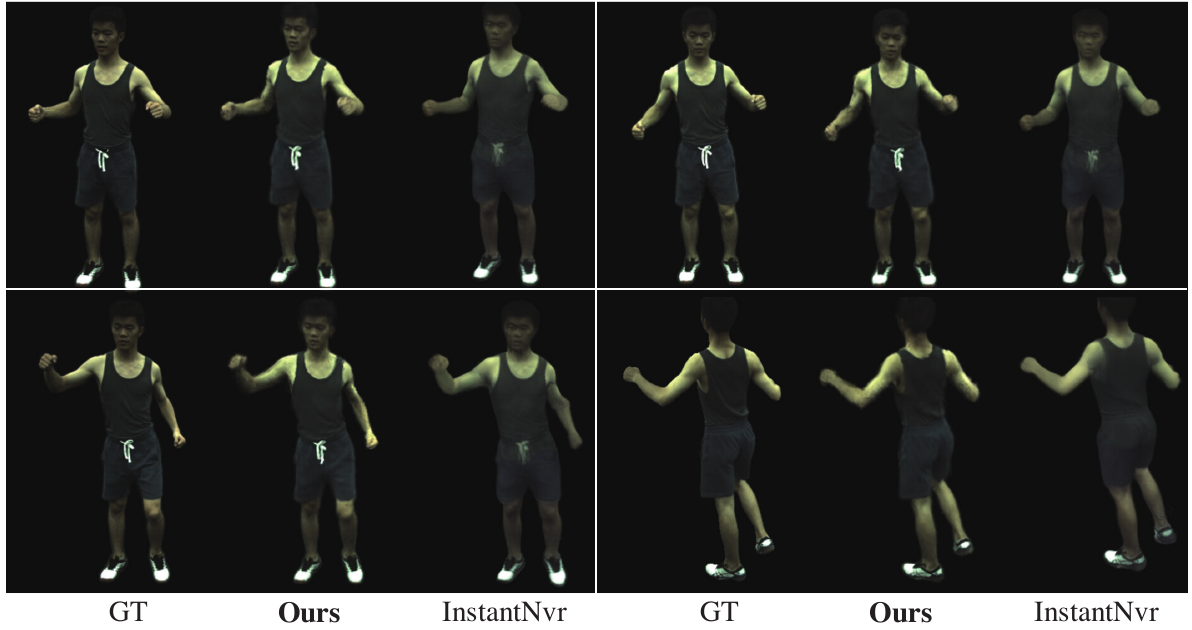}
    \captionsetup{type=figure}
    \vspace{-1.5em}
    \caption{\textbf{Multi view results.} Qualitative results of methods trained with 4 views on the Sequence 377 of the ZJU-MoCap dataset. }
    \label{fig:sup_multi_view}
\end{figure*}

%% file: figs/tex/sup_novel_pose.tex
\begin{figure*}
    \centering
    \includegraphics[width=\linewidth]{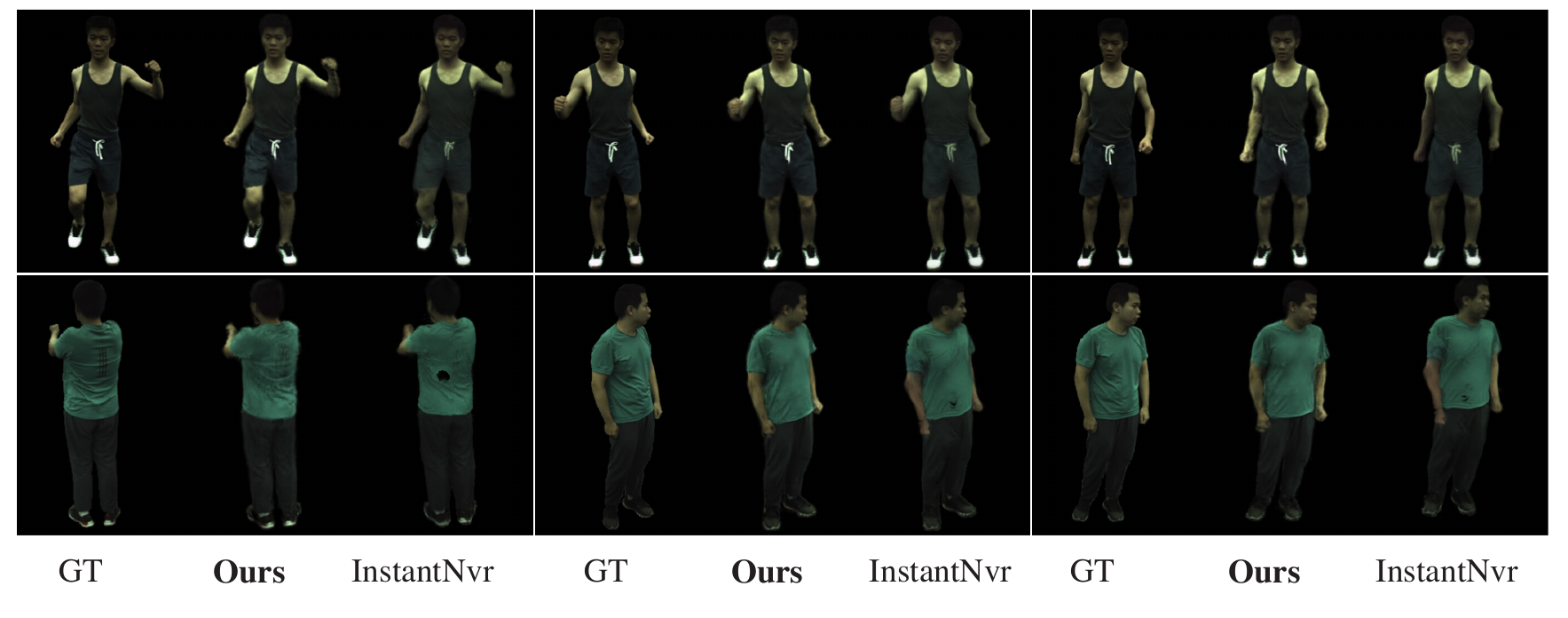}
    \captionsetup{type=figure}
    \caption{\textbf{Novel pose results. (\S~\ref{sec:sup_novelpose})} We show the results of unseen poses of our model and~\cite{instant_nvr}. Results show that our model is less likely to produce artifacts or holes in unseen pose synthesis. }
    \label{fig:sup_novel_pose}
\end{figure*}

%% file: tables/novel_view_1024_sepa.tex
\begin{table*}
    \centering
    \footnotesize
    \begin{tabular}{r|c|cccc|cccc}
    \Xhline{2\arrayrulewidth}
    \rowcolor[HTML]{C0C0C0} 
\multicolumn{10}{c}{ZJU-MoCap~\cite{peng2021neuralbody}}\\
 \Xhline{2\arrayrulewidth}
 \rowcolor[HTML]{C0C0C0} 
 Method& Training Time& PSNR& SSIM& LPIPS* & FPS & PSNR& SSIM& LPIPS* &FPS \\ 
 \Xhline{2\arrayrulewidth}
  \multicolumn{2}{r|}{Subject}&\multicolumn{4}{c|}{377}&\multicolumn{4}{c}{386}\\ \hline
        3D GS~\cite{kerbl3Dgaussians}&  5min&  26.17&  0.949&  60.96& 156& 30.17& 0.951& 51.81&156 
        \\ 
        InstantNvr~\cite{instant_nvr}&  5min&  31.69&  0.981&  32.04& 1.53 & 33.16& 0.979&38.67&1.53\\ 
        InstartAvatar~\cite{Jiang_2023_CVPR_instantavatar}&  5min&  29.90 &  0.961&  49.00&  8.75& 30.67& 0.917& 111.5 & 8.75 \\
 \hline
 Ours& 100s & 32.18& 0.977& 24.65& 104& 33.94& 0.972& 36.03&104\\
 Ours& 5min& 32.02& 0.976& 21.35& 104& 33.78& 0.969& 33.73&104\\ 
 \hline
         \multicolumn{2}{r|}{Subject}&\multicolumn{4}{c|}{387}&\multicolumn{4}{c}{392}\\ \hline
        3D GS~\cite{kerbl3Dgaussians}& 5min& 24.56& 0.922& 80.61&156 
        & 26.72& 0.932& 79.61& 156
        \\ 
        InstantNvr~\cite{instant_nvr}& 5min& 27.73& 0.961& 55.90& 1.53& 31.81& 0.973& 39.25& 1.53
        \\ 
        InstartAvatar~\cite{Jiang_2023_CVPR_instantavatar}& 5min& 27.49 & 0.928 & 86.30 & 8.75& 29.39 & 0.934 & 96.90 & 8.75
        \\
 \hline
 Ours& 100s & 28.32& 0.956& 47.76& 104& 32.22& 0.966& 41.89&104\\
 Ours& 5min& 28.26& 0.956& 44.57& 104& 32.11& 0.967& 39.23&104\\ 
 \hline
         \multicolumn{2}{r|}{Subject}&\multicolumn{4}{c|}{393}&\multicolumn{4}{c}{394}\\ \hline
        3D GS~\cite{kerbl3Dgaussians}& 5min
        & 25.01& 0.923& 85.80& 156
        & 26.79& 0.932& 71.38&156 \\ 
        InstantNvr~\cite{instant_nvr}& 5min& 29.46& 0.964& 46.68& 1.53
        & 31.26& 0.969& 39.89
        &1.53 \\ 
        InstartAvatar~\cite{Jiang_2023_CVPR_instantavatar}& 5min& 28.17&  0.931&  86.60 & 8.75
        & 29.64& 0.943&64.20 & 8.75
        \\
 \hline
 Ours& 100s & 29.69& 0.957& 46.52& 104& 31.37& 0.967& 40.16&104\\
 Ours& 5min& 29.52& 0.956& 44.15& 104& 31.25& 0.968& 36.86&104\\  \Xhline{2\arrayrulewidth}
 \rowcolor[HTML]{C0C0C0} 
\multicolumn{10}{c}{MonoCap~\cite{peng2021animatablenerf}}\\\Xhline{2\arrayrulewidth}
 \rowcolor[HTML]{C0C0C0} 
 Method& Training Time& PSNR& SSIM& LPIPS* & FPS & PSNR& SSIM& LPIPS* &FPS \\ 
\Xhline{2\arrayrulewidth}
          \multicolumn{2}{r|}{Subject}&\multicolumn{4}{c|}{Lan}&\multicolumn{4}{c}{Marc}\\ \hline
         3D GS~\cite{kerbl3Dgaussians}&  5min&  28.76&  0.970&   30.19& 156& 30.16& 0.972& 30.76&156 \\ 
         InstantNvr~\cite{instant_nvr}&  5min&  32.78&  0.987&   17.13& 1.53& 33.84& 0.989& 16.92&1.53 \\ 
         InstartAvatar~\cite{Jiang_2023_CVPR_instantavatar}&  5min&  32.43 &  0.978 &  20.90 & 8.75& 33.88 & 0.979 & 24.40&8.75 \\
 \hline
 Ours& 100s & 32.63& 0.982& 14.21& 104& 34.84& 0.983& 19.21&104\\
 Ours& 5min& 32.56& 0.982& 13.20& 104& 35.02& 0.983& 17.25&104\\ 
 \hline
         \multicolumn{2}{r|}{Subject}&\multicolumn{4}{c|}{Olek}&\multicolumn{4}{c}{Vlad}\\ \hline
        3D GS~\cite{kerbl3Dgaussians}& 5min& 28.32& 0.961&  45.24& 147& 23.13& 0.961&  51.16&147 \\ 
        InstantNvr~\cite{instant_nvr}& 5min& 34.95&  0.991& 13.93& 1.48&28.88& 0.984&  18.72&1.48 \\ 
        InstartAvatar~\cite{Jiang_2023_CVPR_instantavatar}& 5min& 34.21 & 0.980 & 20.60 &8.43 & 28.20 & 0.972 & 34.00 & 8.43\\
 \hline
 Ours& 100s & 34.31& 0.982& 15.07& 101& 28.96& 0.977& 23.56&101\\
 Ours& 5min& 34.09& 0.983& 14.09& 101& 28.84& 0.977& 21.49&101\\  \Xhline{2\arrayrulewidth}
    \end{tabular}
    \caption{512 \( \times \) 512 results of each subject on ZJU-MoCap dataset and Monocap dataset for \textbf{novel view synthesis}  (\S~\ref{sec:sup_novelview_quan}).
    }
    \label{tab:sup_novel_view_512}
    \vspace{2.0em}
\end{table*}

%% file: tables/novel_view_512_sepa.tex
\begin{table*}
    \centering
    \footnotesize
\begin{tabular}{r|c|cccc|cccc}
     
    \Xhline{2\arrayrulewidth}
    \rowcolor[HTML]{C0C0C0} 
\multicolumn{10}{c}{ZJU-MoCap~\cite{peng2021neuralbody}}\\
 \rowcolor[HTML]{C0C0C0} 
 \Xhline{2\arrayrulewidth}
 Method& Training Time& PSNR& SSIM& LPIPS* & FPS & PSNR& SSIM& LPIPS* &FPS \\ 
 \Xhline{2\arrayrulewidth}
  \multicolumn{2}{r|}{Subject}&\multicolumn{4}{c|}{377}&\multicolumn{4}{c}{386}\\ \hline
        3D GS~\cite{kerbl3Dgaussians}&  5min&  26.03&  0.957&  50.22&51.3 & 30.17& 0.958& 47.97& 51.3
        \\ 
        InstantNvr~\cite{instant_nvr}&  5min&  31.69&  0.981&  32.04& 0.5 & 33.16& 0.979&38.67&0.5\\ 
        InstartAvatar~\cite{Jiang_2023_CVPR_instantavatar}&  5min&  27.74&  0.933&  87.91 &  3.83& 28.81 & 0.916 & 97.72 &3.83 \\
 \hline
 Ours& 100s & 31.76& 0.977& 30.27& 68& 33.66& 0.973& 37.30&68\\
 Ours& 5min& 31.64& 0.976& 27.99& 68& 33.42& 0.973& 36.03&68\\ 
 \hline
         \multicolumn{2}{r|}{Subject}&\multicolumn{4}{c|}{387}&\multicolumn{4}{c}{392}\\ \hline
        3D GS~\cite{kerbl3Dgaussians}& 5min& 24.57& 0.931& 64.75& 51.3
        & 26.69& 0.9432& 60.72& 51.3
        \\ 
        InstantNvr~\cite{instant_nvr}& 5min& 27.93& 0.968& 49.11& 0.5& 31.89& 0.977& 42.49&0.5 
        \\ 
        InstartAvatar~\cite{Jiang_2023_CVPR_instantavatar}& 5min& 26.15& 0.890& 107.7& 3.83& 27.98 & 0.9052 & 106.9& 3.83
        \\
 \hline
 Ours& 100s & 27.95& 0.959& 47.56&68 & 31.97& 0.970& 41.65&68\\
 Ours& 5min& 28.02& 0.960& 46.03& 68& 31.86& 0.969& 40.83&68\\ 
 \hline
         \multicolumn{2}{r|}{Subject}&\multicolumn{4}{c|}{393}&\multicolumn{4}{c}{394}\\ \hline
        3D GS~\cite{kerbl3Dgaussians}& 5min
        & 24.97& 0.932& 67.65& 51.3
        & 26.72& 0.941& 58.07&51.3 \\ 
        InstantNvr~\cite{instant_nvr}& 5min& 29.32& 0.969& 48.36&0.5 
        & 31.36& 0.968&39.58&0.5 
         \\ 
        InstartAvatar~\cite{Jiang_2023_CVPR_instantavatar}& 5min& 27.43 & 0.899 & 102.6 & 3.83
        & 28.62 & 0.926 & 81.20 & 3.83
        \\
 \hline
 Ours& 100s & 29.52& 0.961& 46.08&68 & 31.10& 0.964& 41.39&68\\
 Ours& 5min& 29.42& 0.960& 44.64& 68& 31.04& 0.963& 40.07&68\\  \Xhline{2\arrayrulewidth}
 \rowcolor[HTML]{C0C0C0} 
\multicolumn{10}{c}{MonoCap~\cite{peng2021animatablenerf}}\\\Xhline{2\arrayrulewidth}
 \rowcolor[HTML]{C0C0C0} 
 Method& Training Time& PSNR& SSIM& LPIPS* & FPS & PSNR& SSIM& LPIPS* &FPS \\ 
\Xhline{2\arrayrulewidth}
          \multicolumn{2}{r|}{Subject}&\multicolumn{4}{c|}{Lan}&\multicolumn{4}{c}{Marc}\\ \hline
         3D GS~\cite{kerbl3Dgaussians}&  5min&  28.44&  0.974&   25.95& 51.3& 30.13& 0.9762& 26.66&51.3 \\ 
         InstantNvr~\cite{instant_nvr}&  5min&  32.61& 0.988 & 12.73  & 0.5& 33.76& 0.989 & 17.01 & 0.5\\ 
         InstartAvatar~\cite{Jiang_2023_CVPR_instantavatar}&  5min&  32.89 &  0.982&  17.30 & 3.83& 33.72& 0.982 & 21.81 & 3.83\\
 \hline
 Ours& 100s & 31.77& 0.982& 16.38& 68& 34.43& 0.984& 20.29&68\\
 Ours& 5min& 31.72& 0.982& 15.55& 68& 34.56& 0.985& 18.96&68\\ 
 \hline
         \multicolumn{2}{r|}{Subject}&\multicolumn{4}{c|}{Olek}&\multicolumn{4}{c}{Vlad}\\ \hline
        3D GS~\cite{kerbl3Dgaussians}& 5min& 28.34& 0.966&  33.12& 49.6& 23.14& 0.962&  51.73&49.6 \\ 
        InstantNvr~\cite{instant_nvr}& 5min& OOM&OOM &OOM  &OOM & OOM&OOM &OOM  &OOM \\ 
        InstartAvatar~\cite{Jiang_2023_CVPR_instantavatar}& 5min& 34.10 & 0.983 & 18.10 & 3.42& 28.27 & 0.967 & 42.60 &3.42 \\
 \hline
 Ours& 100s & 34.04& 0.984& 16.19&63 & 28.53& 0.979& 20.37&63\\
 Ours& 5min& 33.85& 0.983& 15.32& 63&28.40 & 0.980&19.11&63\\  \Xhline{2\arrayrulewidth}
    \end{tabular}
    \caption{1024 \( \times \) 1024 results of each subject on ZJU-MoCap dataset and Monocap dataset for \textbf{novel view synthesis (\S~\ref{sec:sup_novelview_quan})}.
    }
    \label{tab:sup_novel_view_1024}
\end{table*}

%% file: tables/novel_pose_1024_sepa.tex

%% file: tables/novel_pose_512_sepa.tex

%% file: tables/novel_view_1024_mutli_view.tex
\begin{table*}
    \centering
   \begin{tabular}{rcccccc}
   \Xhline{2\arrayrulewidth}
\rowcolor[HTML]{C0C0C0} 
Method         & Training Time     & PSNR  & SSIM  & LPIPS* & FPS  \\
\Xhline{2\arrayrulewidth}
NeuralBody     & \( \sim \)10hours & 32.99 & 0.983 & 26.8   & 3.5  \\
HumanNeRF      & \( \sim \)10hours & 32.28 & 0.982 & 19.6   & 0.36 \\
AnimatableNeRF & \( \sim \)10hours & 32.31 & 0.980 & 32.2   & 2.1  \\
AnimatableSDF  & \( \sim \)10hours & 32.63 & 0.983 & 32.0   & 1.3  \\
InstantNvr     & \(\sim \)13mins    & 32.55 & 0.981 & 26.5   & 1.5  \\
\hline
\textbf{Ours}           & \( \sim \) 5mins   & 33.90 & 0.981 & 24.92  & 104  \\
\Xhline{2\arrayrulewidth}
\end{tabular}
    \caption{\textbf{Multi-view results comparison (\S~\ref{sec:sub_multi-view_quan}).} Though our model is not designed for multi-view settings, we do experiments on 4 views of Sequence 377. Our model produces remarkable results using much less time while achieving good visual quality and evaluation metrics and much higher FPS. }
    \label{tab:zjumocap377sepamultiview}
\end{table*}

%% file: tables/efficiency_comparison.tex
\begin{table*}[]
\centering
\begin{tabular}{rcccc}
\Xhline{2\arrayrulewidth}
\rowcolor[HTML]{C0C0C0} 
Method                          & Resolution & Train Memory & Infer Memory & Model Size \\
\Xhline{2\arrayrulewidth}
                                & 512\( \times \)512    & 4542M        & 3964M        & 151M       \\
                                & 1024\( \times \)1024  & 4542M        & 4020M        & 151M       \\
                                & 642\( \times \)470    & 4516M        & 3966M        & 151M       \\
\multirow{-4}{*}{InstantAvatar~\cite{Jiang_2023_CVPR_instantavatar}} & 1285\( \times \)940   & 4654M        & 4038M        & 151M       \\ \hline
                                & 512\( \times \)512    & 19132M       & 4816M        & 3.2G       \\
                                & 1024\( \times \)1024  & 23320M       & 4816M        & 3.2G       \\
                                & 642\( \times \)470    & 21868M       & 7660M        & 3.2G       \\
\multirow{-4}{*}{InstantNvr~\cite{instant_nvr}}    & 1285\( \times \)940   & OOM          & -            & -          \\ \hline
                                & 512\( \times \)512    & 1878M        & 956M         & 12M+292K   \\
                                & 1024\( \times \)1024  & 4146M        & 1842M        & 12M+292K   \\
                                & 642\( \times \)470    & 1932M        & 1008M        & 12M+292K   \\
\multirow{-4}{*}{\textbf{Ours}}          & 1285\( \times \)940   & 4726M        & 2038M        & 12M+292K   \\
\Xhline{2\arrayrulewidth}
\end{tabular}
\caption{\textbf{Memory efficiency comparison (\S~\ref{sec:sup_efficiency_compare}).} For all resolutions in the dataset, we test the memory efficiency by Training GPU memory consumption (``Train Memory''), Inference GPU memory consumption (``Infer Memory''), and the size of the checkpoints (``Model Size''). Results demonstrate that our model utilizes much less GPU memory and disk usage than~\cite{instant_nvr} while maintaining comparable or better visual quality. Note: when inferring, we don't precompute and save Gaussians in target space while we choose to query the network for each frame. This methodological choice significantly reduces the storage requirements and makes it possible for Human101 to apply for more flexible use cases.
}
\label{tab:mem_efficiency_comparison}
\end{table*}

%% file: tables/ablation_sparse_frame.tex
\begin{table}[]
\centering
\begin{tabular}{cccc}
\rowcolor[HTML]{C0C0C0} 
\Xhline{2\arrayrulewidth}
Frame Num & PSNR & SSIM & LPIPS* \\
\Xhline{2\arrayrulewidth}
25        &  31.66    &  0.974    &  24.78      \\
50        &  32.00    &  0.975    &  22.26      \\
100       &  \bestcolor32.18    &  \bestcolor0.977    &  21.32      \\
250       &  32.17    &  \bestcolor0.977    &  \bestcolor19.17      \\
\Xhline{2\arrayrulewidth}
\end{tabular}
    \caption{\textbf{Ablation study on frame number (\S~\ref{sec:sup_ablation}).} Our model still maintains good visual quality using sparse frame inputs even with only 25 images to train.
    }
    \label{tab:ablation_sparse_frame}
\end{table}

%% file: tables/ablation_PE.tex
\begin{table}
    \centering
    \begin{tabular}{cccc}
        \rowcolor[HTML]{C0C0C0} 
        \Xhline{2\arrayrulewidth}
        Method & PSNR & SSIM & LPIPS* \\
        \Xhline{2\arrayrulewidth}
        NoEnc  &    32.13  &    0.976  &    24.47   \\
        GridEnc  &  31.99    &  0.975    &  29.47     \\
        PE(\textbf{Ours})       &  \bestcolor 32.18    &  \bestcolor 0.977    &  \bestcolor 21.32     \\
        \Xhline{2\arrayrulewidth}
    \end{tabular}
    \caption{\textbf{Ablation study on encoding method (\S~\ref{sec:sup_ablation}).} The results demonstrate that the positional encoding method produces better quality than no encoding(``NoEnc'') and grid-encoding (``GridEnc'').
    }
    \label{tab:ablation_PE}
\end{table}

%% file: tables/ablation_shs.tex
\begin{table}[]
\centering
\begin{tabular}{cccc}
    \rowcolor[HTML]{C0C0C0} 
    \Xhline{2\arrayrulewidth}
    Degree  & PSNR & SSIM & LPIPS* \\
    \Xhline{2\arrayrulewidth}
    0 & 31.77     &  0.974    &  24.05      \\
    1 & 32.04     &  0.976    &  22.55      \\
    2 & 32.13     &  0.976    &  21.60      \\
    3(\textbf{Ours}) &  \bestcolor 32.18    &  \bestcolor0.977    &   \bestcolor21.32    \\
    \Xhline{2\arrayrulewidth}
\end{tabular}
\caption{\textbf{Ablation study on the degree of spherical harmonics (\S~\ref{sec:sup_ablation}).} We evaluate the impact of the harmonics' degree on the quality of reconstruction, with the degree of 3 (our chosen configuration) offering a trade-off between reconstruction detail and computational efficiency. 
}
\label{tab:sup_ablation_shs}
\end{table}

%% file: figs/tex/sup_converge_speed.tex
\begin{figure}
    \centering
    \includegraphics[width=\linewidth]{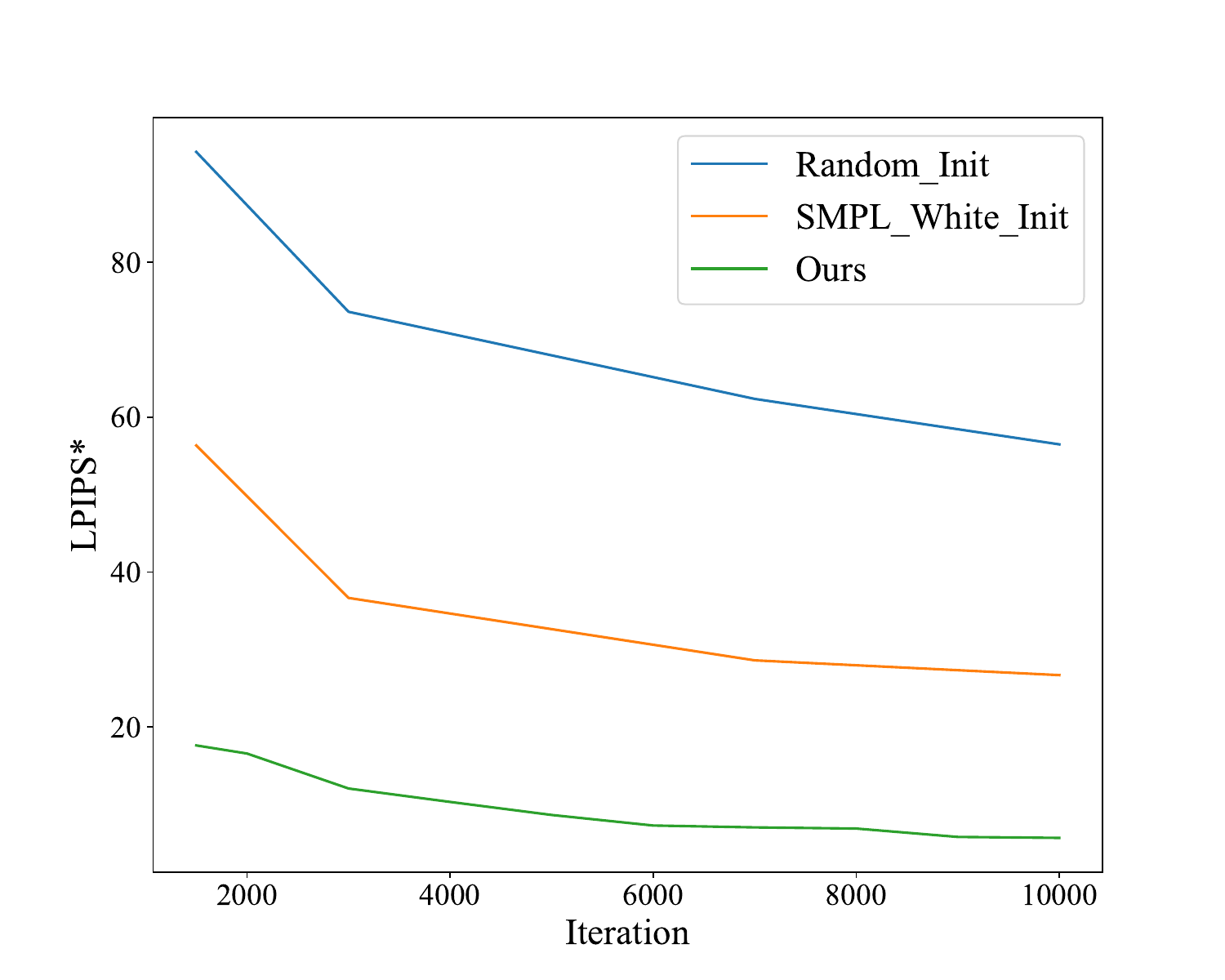}
    \captionsetup{type=figure}
    \vspace{-1.5em}
    \caption{\textbf{Ablation study on convergence speed. (\S~\ref{sec:sup_ablation})} We compare training view LPIPS results with the initialization method to be random initialization (``Random\_Init''), bare SMPL with white color initialization (``SMPL\_White\_Init.'') and our Canonical Human Initialization method (``Ours'') separately. }
    \label{fig:sup_converge_speed}
\end{figure}

%% file: figs/tex/gaussian_count.tex
\begin{figure}
    \centering
    \includegraphics[width=\linewidth]{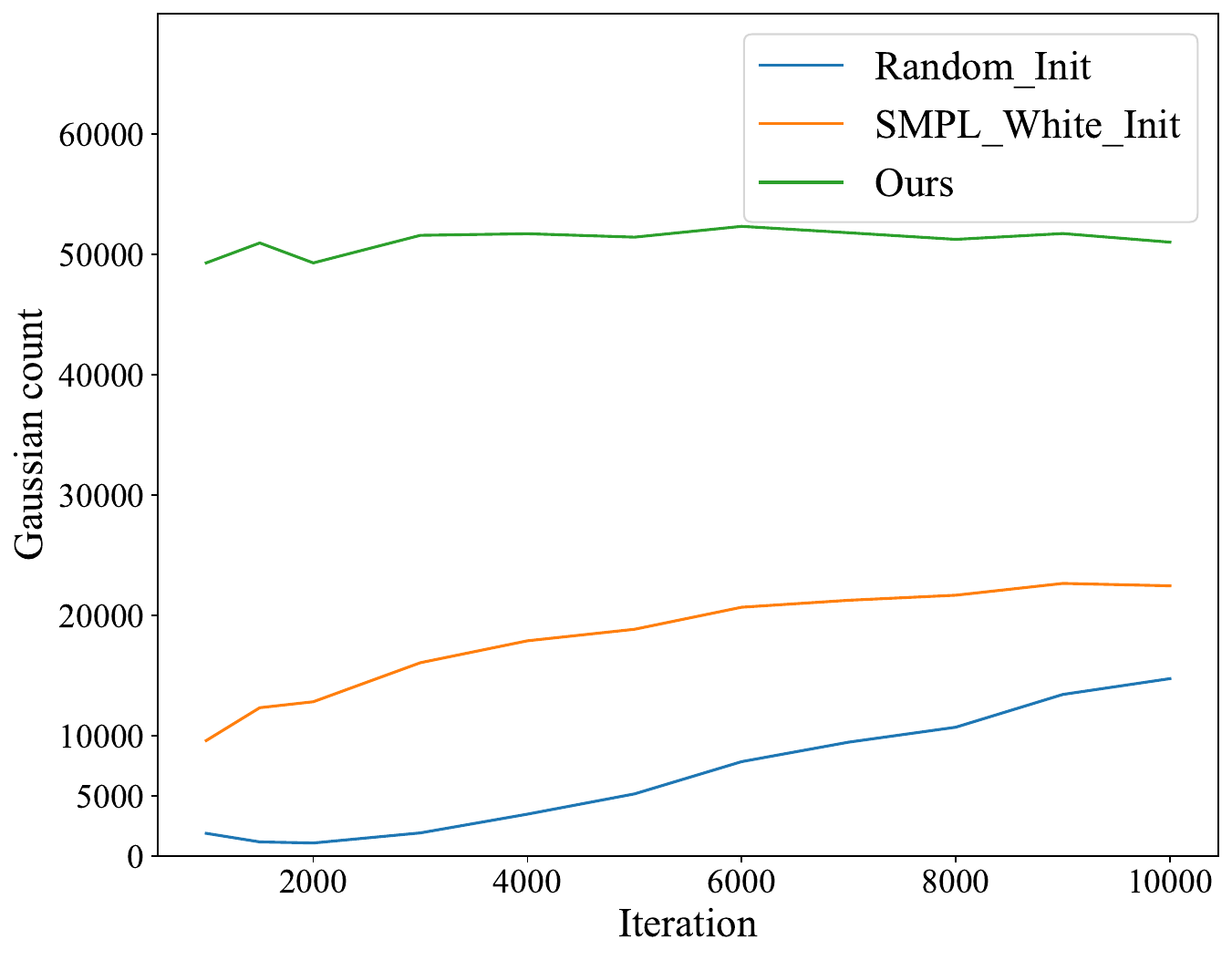}
    \captionsetup{type=figure}
    \vspace{-1.5em}
    \caption{\textbf{Ablation study on Gaussian count (\S~\ref{sec:sup_ablation}).} We compare the number of Gaussians at different stages using various initialization methods. A superior initialization approach necessitates a greater number of Gaussians to represent the geometry more precisely, which generally yields better results. }
    \label{fig:sup_gaussian_count}
\end{figure}

%% file: figs/tex/failure_case.tex
\begin{figure}
    \centering
    \includegraphics[width=0.35\linewidth]{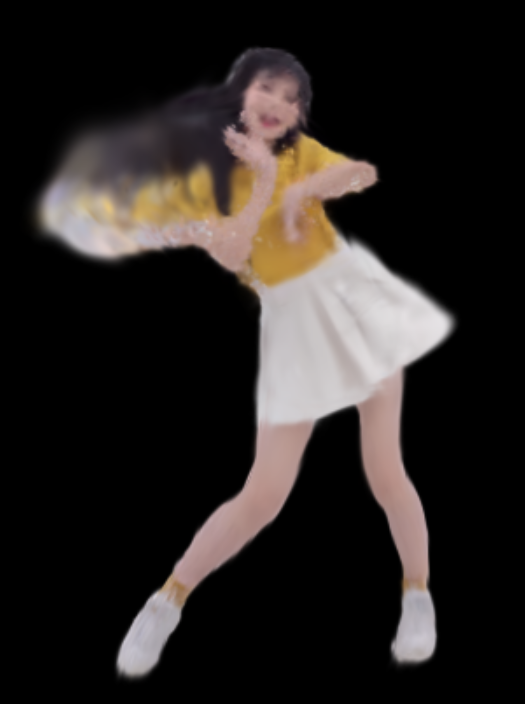}
    \captionsetup{type=figure}
    \caption{\textbf{Failure case (\S~\ref{sec:sup_failure_cases}).} When dealing with intense movement where flowing hair carries unmasked background elements, our model may produce artifacts due to the complex human motion. }
    \label{fig:failure_case}
\end{figure}

%% file: figs/tex/composite.tex
\begin{figure}
    \centering
    \includegraphics[width=0.4\linewidth]{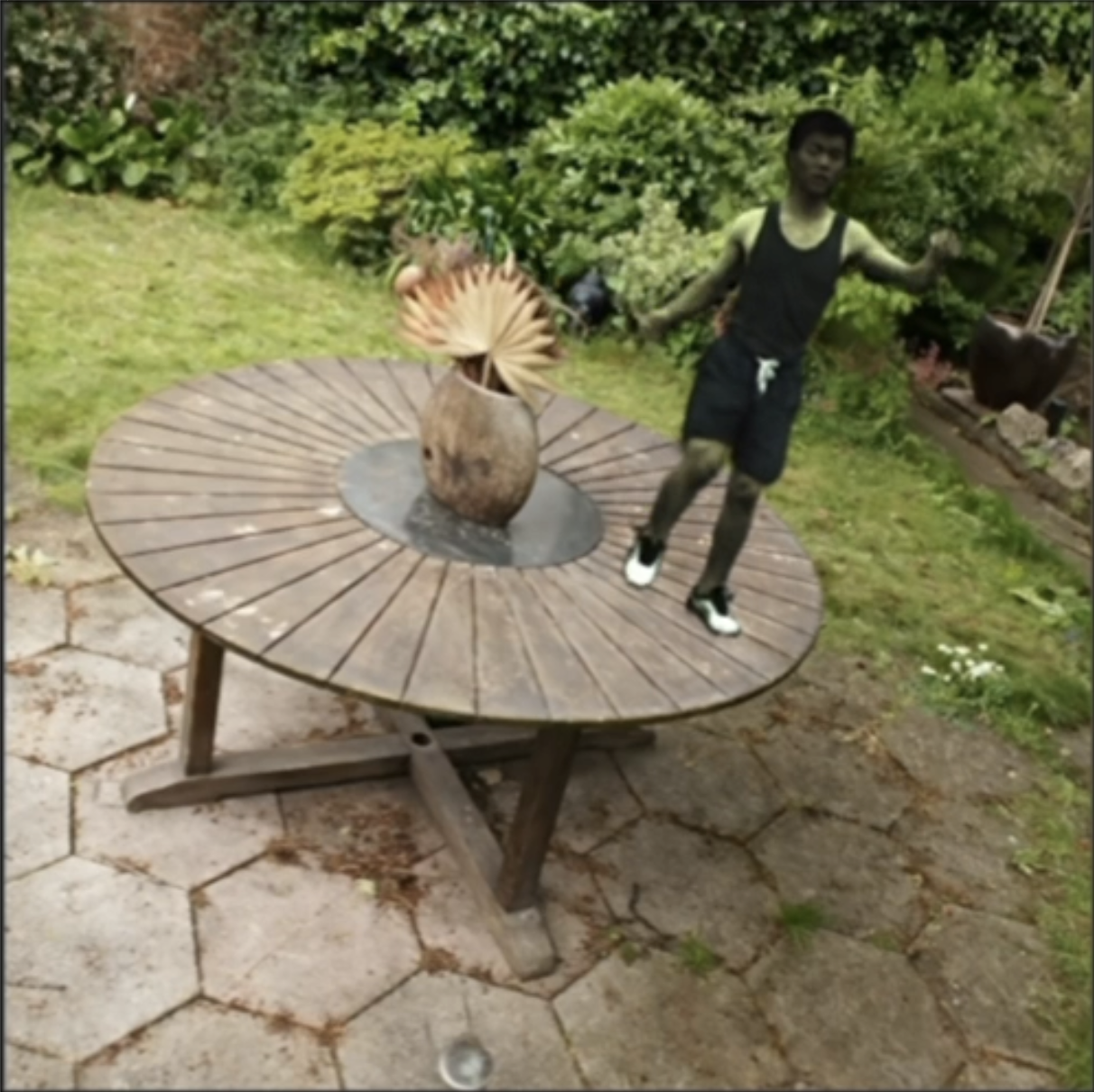}
    \captionsetup{type=figure}
    \caption{\textbf{Composite scene rendering (\S~\ref{sec:sup_composite}).} We render the avatar integrated with the scene. }
    \label{fig:composite}
\end{figure}